\documentclass[11pt]{article}
	
	\newcommand{\blind}{0}
	
    \makeatletter
    \renewcommand\section{\@startsection {section}{1}{\z@}%
                                       {-3.5ex \@plus -1ex \@minus -.2ex}%
                                       {2.3ex \@plus.2ex}%
                                       {\normalfont\fontfamily{phv}\fontsize{16}{19}\bfseries}}
    \renewcommand\subsection{\@startsection{subsection}{2}{\z@}%
                                         {-3.25ex\@plus -1ex \@minus -.2ex}%
                                         {1.5ex \@plus .2ex}%
                                         {\normalfont\fontfamily{phv}\fontsize{14}{17}\bfseries}}
    \renewcommand\subsubsection{\@startsection{subsubsection}{3}{\z@}%
                                        {-3.25ex\@plus -1ex \@minus -.2ex}%
                                         {1.5ex \@plus .2ex}%
                                         {\normalfont\normalsize\fontfamily{phv}\fontsize{14}{17}\selectfont}}
    \makeatother
    
	\usepackage{framed}
	\usepackage{amssymb}
	\usepackage{latexsym}

	\usepackage{url}
	\usepackage{xcolor}

	\usepackage{hyperref}
	\usepackage{textcomp}
	\usepackage{algorithmic}
	\usepackage{algorithm}
	\usepackage{times}
	\usepackage{graphicx}
	\usepackage{amsmath}
	\usepackage{amssymb}
	\usepackage{multirow}
	\usepackage{scrextend}
	\usepackage[inline]{enumitem}
	\usepackage[bold]{hhtensor}
	\usepackage{color}
	\usepackage{float}
	\usepackage{placeins}
	\usepackage[font=normal]{subfig}
	\usepackage{enumitem}
	\usepackage{array}
	\usepackage{subfig}
	\usepackage{mathtools}
	\usepackage{dsfont}
	\usepackage{booktabs}
	\usepackage{makecell}
	\graphicspath{{figures/}}
	
\newcommand{\shorteq}{%
  \settowidth{\@tempdima}{-}
  \resizebox{\@tempdima}{\height}{=}%
}
\newcommand{\PreserveBackslash}[1]{\let\temp=\\#1\let\\=\temp}
\newcommand{\pluseq}{\mathrel{+}=}

\newcolumntype{C}[1]{>{\PreserveBackslash\centering}p{#1}}
\newcolumntype{L}[1]{>{\PreserveBackslash}p{#1}}
\newcommand*\samethanks[1][\value{footnote}]{\footnotemark[#1]}

\setenumerate[1]{itemsep=1.5pt,partopsep=0pt,parsep=\parskip,topsep=5pt}
\setitemize[1]{itemsep=1.5pt,partopsep=0pt,parsep=\parskip,topsep=5pt}
\setdescription{itemsep=0pt,partopsep=0pt,parsep=\parskip,topsep=5pt}

\definecolor{newcolor}{rgb}{.8,.349,.1}

\begin{document}
		
		\def\spacingset#1{\renewcommand{\baselinestretch}%
			{#1}\small\normalsize} \spacingset{1}
		
		\if0\blind
		{
			\title{\bf SSMD: Semi-Supervised Medical Image Detection with Adaptive Consistency and Heterogeneous Perturbation}
			\author{Hong-Yu Zhou$^{a,b,}$\thanks{First three author contributed equally.}\ , Chengdi Wang$^{a,}$\samethanks\ , Haofeng Li$^{c,}$\samethanks\ , Gang Wang$^a$, Shu Zhang$^d$, \\ Weimin Li$^{a,}$\thanks{Corresponding author.}\ , Yizhou Yu$^{b,}$\samethanks\ \vspace{3mm}\\
			$^a$ Department of Respiratory and Critical Care Medicine, National Clinical Research\\ Center for Geriatrics, Frontiers Science Center for Disease-related Molecular Network,\\ West China Hospital, Sichuan University, Chengdu 610041, P.R. China \vspace{1.5mm}\\
             $^b$ Department of Computer Science, The University of Hong Kong,\\ Pokfulam, Hong Kong\vspace{1.5mm}\\
             $^c$ Shenzhen Research Institute of Big Data, The Chinese University of\\ Hong Kong, Shenzhen 518000, P.R. China\vspace{1.5mm}\\
			 $^d$ AI Lab, Deepwise Healthcare, Beijing 100080, P.R. China}
			\date{}
			\maketitle
		} \fi
		
		\if1\blind
		{

            \title{\bf \emph{IISE Transactions} \LaTeX \ Template}
			\author{Author information is purposely removed for double-blind review}
			
\bigskip
			\bigskip
			\bigskip
			\begin{center}
				{\LARGE\bf \emph{IISE Transactions} \LaTeX \ Template}
			\end{center}
			\medskip
		} \fi
		\bigskip
		
	\begin{abstract}
Semi-Supervised classification and segmentation methods have been widely investigated in medical image analysis. Both approaches can improve the performance of fully-supervised methods with additional unlabeled data. However, as a fundamental task, semi-supervised object detection has not gained enough attention in the field of medical image analysis. In this paper, we propose a novel Semi-Supervised Medical image Detector (SSMD). The motivation behind SSMD is to provide free yet effective supervision for unlabeled data, by regularizing the predictions at each position to be consistent.
To achieve the above idea, we develop a novel adaptive consistency cost function to regularize different components in the predictions. Moreover, we introduce heterogeneous perturbation strategies that work in both feature space and image space, so that the proposed detector is promising to produce powerful image representations and robust predictions. Extensive experimental results show that the proposed SSMD achieves the state-of-the-art performance at a wide range of settings. We also demonstrate the strength of each proposed module with comprehensive ablation studies.
	\end{abstract}
			
	\noindent%

	\spacingset{1.5} 

\section{Introduction}
\begin{figure*}[t]
	\centering
	\includegraphics[width=0.75\textwidth]{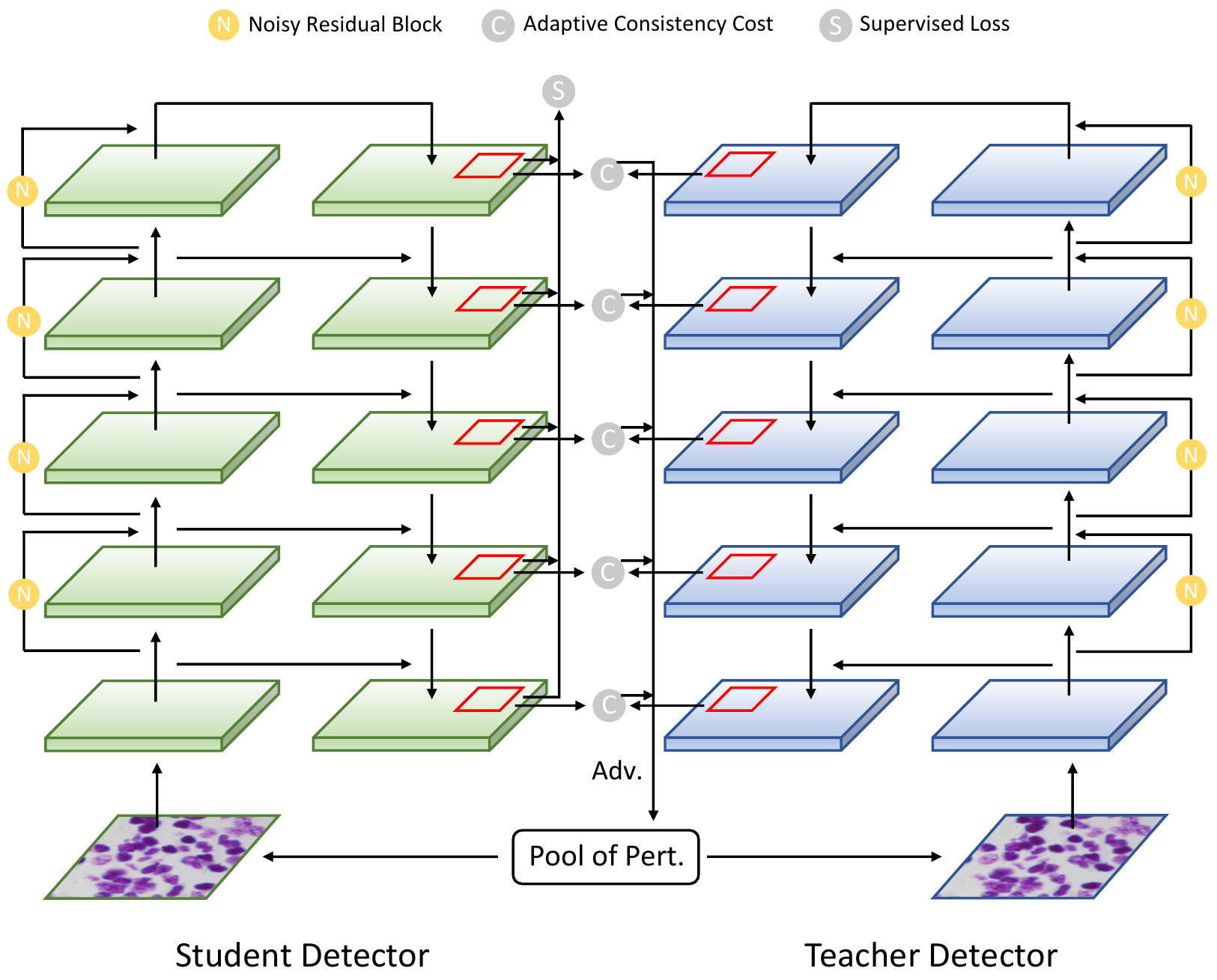}
	\caption{Overview of the proposed Semi-Supervised Medical image Detector (SSMD). Two feature pyramid networks are utilized to predict consistent outputs. \textbf{Pool of Pert.} refers to a set of perturbation strategies, which include horizontal flip, vertical flip, random rotation and adversarial perturbation (denoted as \textbf{Adv.}). 
	Note that the adversarial perturbation is only applied to the input of teacher network.}
	\label{framework}
\end{figure*}

Recently, deep convolution neural networks have achieved remarkable success in processing and understanding visual data, the convolution layers with learnable parameters in CNNs can adaptively harvest powerful image representations based on a large manually-annotated dataset. Sometimes collecting and labeling such a large-scale dataset can be expensive, time-consuming and unaffordable in real applications. Particularly, annotating medical images usually require well-trained experts with prior biomedical knowledge. On the other hand, it is easier to obtain unlabeled data. Thus, how to train neural networks with both labeled data and unlabeled ones becomes an important problem, and also refers to the setting of semi-supervised learning.

Nowadays, a lot of deep semi-supervised methods~\cite{berthelot2019mixmatch,laine2016temporal,miyato2018virtual,tarvainen2017mean,lee2013pseudo} have been developed. They are effective in leveraging unlabeled data to mitigate the dependence of deep learning models on large-scale annotated datasets. 
In the field of medical image analysis, numerous semi-supervised learning methods have been proposed for a wide-range of applications, such as abnormality classification \cite{lecouat2018semi}, 2D image segmentation \cite{li2018semi,zhang2017deep} and 3D volume segmentation \cite{chen2019multi,nie2018asdnet,yu2019uncertainty,li2018semi}. However, the community seldomly investigates the object detection task for medical images from a perspective of semi-supervised learning.
{In this paper, we mainly focus on two object detection tasks: lesions detection and nuclei detection. Locating lesions and abnormalities in CT scans is a primary object detection task for radiologists. They need to find out the location of lesions, and describe the related attributes in radiological reports. An automatic lesion detector could not only reduce the workload of radiologists, but also benefit the areas that have a shortage of experienced radiologists. DeepLesion, a representative lesion detection dataset, contains 32,120 CT slices and 32,735 annotated lesions. Another fundamental object detection task in medical image analysis is nuclei or cell detection, which helps measure quantitative information to better understand disease progression. 2018 Data Science Bowl introduced a nuclei detection dataset that consists of about 27,000 cells. To reduce the high cost of medical image annotations, it is important to train a robust detector with not only labeled but also unlabeled medical images.}
Thus, in this paper we propose a novel Semi-Supervised Medical image Detector (SSMD) that can make use of unlabeled medical images to produce robust representations in an effective way.

Consistency-based semi-supervised learning methods \cite{jeong2019consistency} mainly utilize self-supervision \cite{zhou2019models,chen2019self,zhou2020C2L} and usually consist of two procedures. First, synthesize a pair of input images via some data augmentation strategies. Second, force the paired model's outputs to be consistent, and formulate such constraint as an additional loss to train neural networks.
The assumption behind is that deep learning models should produce similar representations or predictions with these augmented inputs. Therefore, such consistency constraint can be applied to the model outputs corresponding to the augmented inputs, and serves as an additional supervision when the ground-truth annotations are not available. However, some existing semi-supervised object detectors only adopt simple data augmentations such as image flipping, and do not consider the confidence of proposals in the consistency-based loss of semi-supervised detection.

Our proposed semi-supervised detector addresses two problems when applying the consistency regularization to medical image detection: a) too many background proposals may dominate the training procedure and b) mediocre augmentation strategies (such as horizontal flip and translation) cannot well regularize visual representations. For issue a), we propose \emph{adaptive consistency cost} to adaptively scale the loss values, where the scaling factor decays to zero as confidence of the background class increases. Intuitively, the proposed adaptive mechanism can automatically down-weight the influences of background proposals. As for problem b), we introduce a set of heterogeneous perturbation methods to advance the regularization effect of the consistency loss. The core idea behind is that we want the detector to capture the invariant representations as we apply various perturbations. We believe these representations are more robust and thus more generalized, even under various real-world noise.

\section{Related Work}
Many methods have been proposed to solve semi-supervised learning (SSL) problems. Here we mainly focus on deep learning based approaches which are  most related to our work. 
Recent studies in semi-supervised learning could be categorized into two groups: pseudo labels and consistency regularization. In this section, we review these two types of methods from three different aspects: natural image classification, medical image analysis and semi-supervised object detection.

\subsection{SSL in Natural Image Classification}
\cite{lee2013pseudo} is the first to introduce pseudo-labeling for deep neural networks. Pseudo-labeling first trains a model with labeled data, predicts classification probability on unlabeled data, and then annotate the class with highest probability as pseudo labels. Lastly, all the data with real or pseudo labels are utilized to retrain the classification model.
To obtain more accurate pseudo labels, \cite{berthelot2019mixmatch} proposed MixMatch which employs a sharpen function to average the predictions of stochastic augmented inputs. In MixMatch, Mixup \cite{zhang2017mixup} was utilized as a strong augmentation method that could increase the diversity of both labeled and unlabeled data. 
{\cite{berthelot2019remixmatch} improved MixMatch by tackling the distribution misalignment and introducing more augmentation strategies.} \cite{wang2019repetitive} provided a theoretical explanation for pseudo-labeling.

Besides from pseudo-labeling methods, \cite{laine2016temporal} proposed $\pi$-model which encourages consistent network outputs between two realizations of the same input stimulus. Such consistency works as a supervision for unlabeled images, and is easily incorporated into training loss. \cite{tarvainen2017mean} developed Mean Teacher that takes an exponential moving average of model weights to obtain more accurate predictions for consistency regularization. Different from Mean Teacher, \cite{ke2019dual} decoupled the connection between student and teacher networks, and used another student network to substitute for the teacher model. \cite{verma2019interpolation} proposed Interpolation Consistency Training which enforces the prediction at an interpolation of unlabeled points to match with the interpolation of the predictions at those points.

The most related work to ours is Virtual Adversarial Training \cite{miyato2018virtual} which improves the robustness of the conditional label distribution around each input data point against local perturbation. However, our proposed detector is not simply transferring adversarial loss to semi-supervised classification. First, our proposed method makes use of position information to synthesize more effective adversarial samples. Second, the proposed adversarial perturbation considers the influence of different instances. Furthermore, we explore more perturbation strategies in both feature space and image space.

\subsection{SSL in Medical Image Analysis}
SSL methods are commonly used in medical image analysis to address the lack of manually annotated data. \cite{lecouat2018semi} proposed a patch-based semi-supervised learning approach and applied it to the classification of diabetic retinopathy from funduscopic images. \cite{madani2018semi}, \cite{madani2018deep} and \cite{yi2018unsupervised} used generative adversarial networks to conduct semi-supervised classification in chest X-ray, cardiology and dermoscopy, respectively. Recently, \cite{zhou2019collaborative} developed a collaborative learning method to jointly improve the performance of disease grading and lesion segmentation, via an attention-based semi-supervised learning mechanism. \cite{liu2020semi} exploited unlabeled data by modeling the relation consistency among different samples, rather than only enforcing individual consistency.

Apart from image classification problems, SSL has also been applied to medical image segmentation task. \cite{chen2019multi} presented a new multi-task attention-based segmentation framework by enforcing consistency regularization on reconstructed foreground and background. \cite{yu2019uncertainty} and \cite{li2020transformation} introduced a student-teacher framework which employs prediction uncertainty to highlight reliable consistent predictions. A soft-label based semi-supervised segmentation approach was presented in \cite{chang2020soft} to improve the ventricle segmentation of 2D cine MR images.

However, these above works do not aim at the object detection problem for medical images. To fill in such a gap, in this paper we develop a novel semi-supervised medical detector that emphasizes the importance of producing consistent and robust predictions. 

\subsection{Semi-Supervised Object Detection}
\cite{jeong2019consistency} presented a Consistency-based Semi-Supervised learning  for Object Detection (CSD) that works well for both single-stage and two-stage detectors.
Compared with the consistency regularization in semi-supervised classification, the proposed consistency constraint in CSD is applied to not only object classification but also object localization of a predicted region. Lately, following MixMatch \cite{berthelot2019mixmatch}, \cite{wang2020focalmix} built a novel semi-supervised lesion detector FocalMix based on Mixup \cite{zhang2017mixup} and a soft-version focal loss \cite{lin2017focal}.
{\cite{sohn2020simple} proposed a semi-supervised learning object detection framework, STAC, which is based on high-confidence pseudo labels and the consistency via data augmentations. Different from STAC, our proposed method achieves different and stronger data augmentations by introducing Gaussian and adversarial noises to feature and image spaces.}

In this paper, our proposed object detector is built on top of consistency regularization and CSD, instead of pseudo-labeling in FocalMix. 
Compared with CSD, the proposed approach employs heterogeneous perturbations to enhance the robustness of predictions as well as the detection accuracy. In addition, we develop a novel adaptive cost function to model instance-level consistency. 

\begin{algorithm*}[t]
	\caption{Procedure of Semi-Supervised Medical Detection} 
	\label{procedure} 
	\begin{algorithmic}[1]
		\REQUIRE ~~\\
		A batch of labeled images $\mathcal{X}=\{x^1,...,x^B\}$. A batch of unlabeled images $\mathcal{\tilde{X}}=\{\tilde{x}^1,...,\tilde{x}^B\}$. {\color{gray} $\backslash\backslash \ B$ is the batch size} \\
		$f_{\theta^n_s}(\cdot)$: Student network at time step $n$. $f_{\theta^n_t}(\cdot)$: Teacher network at time step $n$.\\
		\STATE $\mathcal{X}_s$=cutout(Rot.($\mathcal{X}$)); $\mathcal{\tilde{X}}_s$=cutout(Rot.($\mathcal{\tilde{X}}$))  {\color{gray} $\backslash\backslash$ Augmentation for the student network}
		\STATE $\mathcal{X}_t$=Adv.(cutout(Rot.(Flip($\mathcal{X}$)))); $\mathcal{\tilde{X}}_t$=Adv.(cutout(Rot.(Flip($\mathcal{\tilde{X}}$))))  {\color{gray} $\backslash\backslash$ Augmentation for the teacher network}
		\STATE {\color{gray} $\backslash\backslash$ Forward pass to get predictions}\\
		\STATE \textbf{for} $n=1$ to $N$ \textbf{do} {\color{gray} $\backslash\backslash\ N$ is number of training iterations}
		\STATE \ \ \ \ \ \ $\theta^{n}_t = \alpha\theta^{n-1}_{t} + (1-\alpha)\theta^{n}_{s}$ {\color{gray} $\backslash\backslash$ Update teacher network}\\
		\STATE \ \ \ \ \ \ $\rm{loss}_{cont}=0$; $\rm{loss}_{sup}=0$ {\color{gray} $\backslash\backslash$ Initialize loss values}
		\STATE \ \ \ \ \ \ \textbf{for} $i=1$ to $B$ \textbf{do} \\
		\STATE \ \ \ \ \ \ \ \ \ \ \ \ $p^c_s, p^x_s, p^y_s, p^w_s, p^h_s = f_{\theta^n_s}(x_s^i)$; $\tilde{p}^c_s, \tilde{p}^x_s, \tilde{p}^y_s, \tilde{p}^w_s, \tilde{p}^h_s = f_{\theta^n_s}(\tilde{x}_s^i)$ {\color{gray} $\backslash\backslash$ Forward through student network}
		\STATE \ \ \ \ \ \ \ \ \ \ \ \ $p^c_t, p^x_t, p^y_t, p^w_t, p^h_t = f_{\theta^n_t}(x_t^i)$; $\tilde{p}^c_t, \tilde{p}^x_t, \tilde{p}^y_t, \tilde{p}^w_t, \tilde{p}^h_t = f_{\theta^n_t}(\tilde{x}_t^i)$ {\color{gray} $\backslash\backslash$ Forward through teacher network}
		
		\STATE \ \ \ \ \ \ \ \ \ \ \ \ $\text{loss}_{sup}\pluseq\text{CE}(p_s^c, p^c_{gt})+\text{SmoothL1}(p^{\{x,y,w,h\}}_s, p^{\{x,y,w,h\}}_{gt})$ {\color{gray} $\backslash\backslash$ Apply supervised loss}
		
		\STATE \ \ \ \ \ \ \ \ \ \ \ \ $\text{loss}_{cont}\pluseq\text{W}(p^c_s, p^c_t)\otimes(\text{KL}(p^c_s, p^c_t)+\text{MSE}(p^{\{x,y,w,h\}}_s, p^{\{x,y,w,h\}}_t))$ {\color{gray} $\backslash\backslash$ Consistency loss on labeled images}\label{loss_cont}
		\STATE \ \ \ \ \ \ \ \ \ \ \ \ $\text{loss}_{cont}\pluseq\text{W}(\tilde{p}^c_s, \tilde{p}^c_t)\otimes(\text{KL}(\tilde{p}^c_s, \tilde{p}^c_t)+\text{MSE}(\tilde{p}^{\{x,y,w,h\}}_s, \tilde{p}^{\{x,y,w,h\}}_t))$ {\color{gray} $\backslash\backslash$ Consistency loss on unlabeled images}
		
		\STATE \ \ \ \ \ \ $\text{loss} = \text{loss}_{sup} + \lambda\text{loss}_{cont}$ {\color{gray} $\backslash\backslash$ $\lambda$ is a hyperparameter} \label{lambda}\\
		\STATE \ \ \ \ \ \ Backward(loss) {\color{gray} $\backslash\backslash$ Update student network}
	\end{algorithmic}
\end{algorithm*}

\section{Method}
\begin{figure}
	\centering 
	\includegraphics[width=0.9\columnwidth]{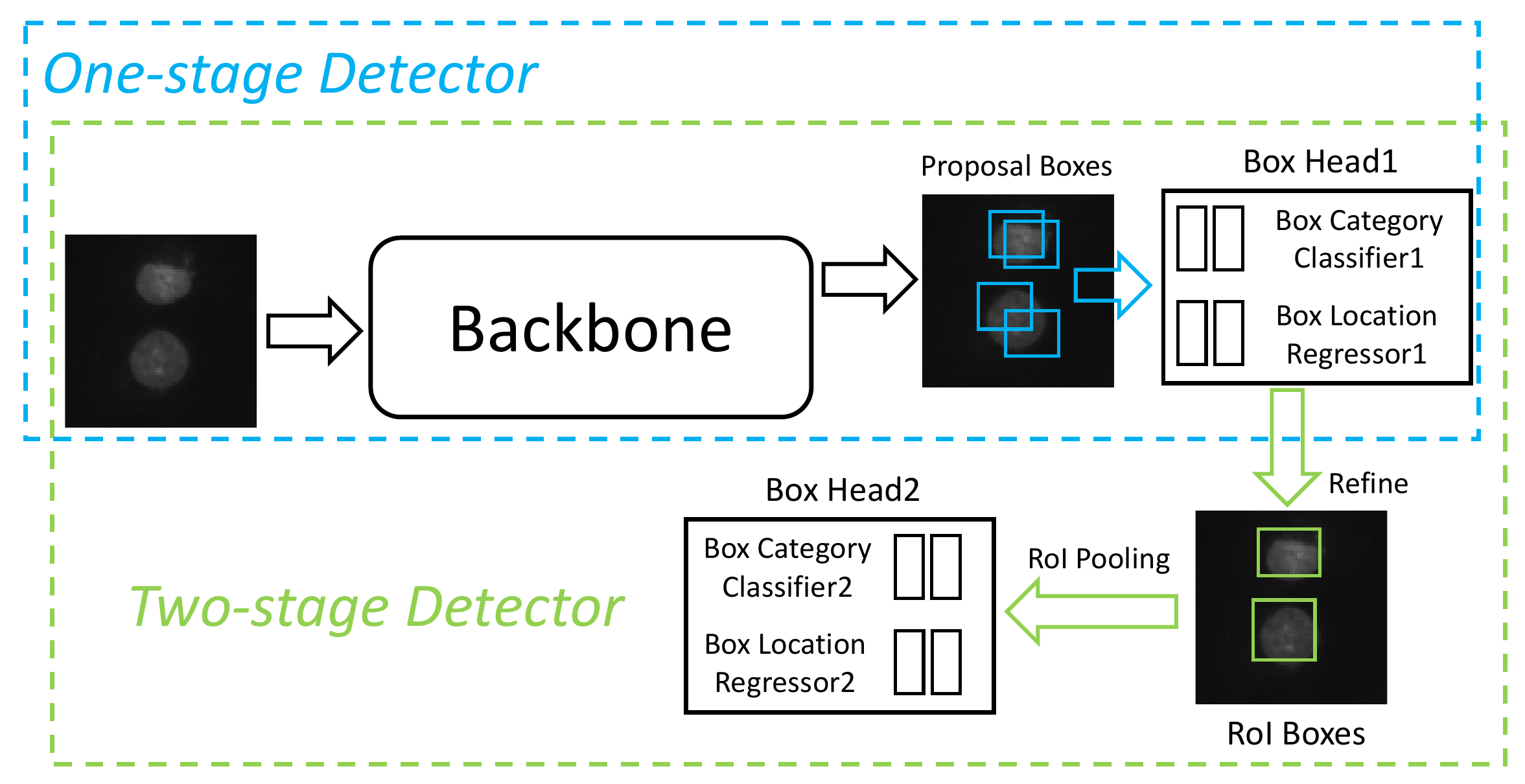}
	\caption{General network architecture for medical image detection. RoI is an abbreviation for region of interest.}
	\label{overview_detection}
\end{figure}
{Existing deep learning based medical image detection methods are usually anchor-based, which predict the relative position and scale factors between each object box and some pre-defined anchor boxes. These methods could be further divided into two types: one-stage \cite{liao2019evaluate,khosravan2018s4nd,zlocha2019improving} and two-stage detectors \cite{wang2020focalmix,ding2017accurate,zhu2018deeplung,pezeshk20183,liu20193dfpn,li2019mvp}, as shown in Figure \ref{overview_detection}. Given a one-stage detector, it first produces a large number of proposal boxes via the backbone network, after which one category classifier and one location regressor are employed to deal with these boxes. In contrast, the two-stage pipeline requires one more box head which is responsible for refining those RoI boxes produced by the first box head. During the training stage, the overall loss function can be summarized as:
\begin{align}
	\begin{split}
		\text{CE}(p^c, p^c_{gt})+\text{SmoothL1}(p^{\{x,y,w,h\}}_s, p^{\{x,y,w,h\}}_{gt}),
	\end{split}
\end{align}
where $p^c$ stands for the class prediction, $p^c_{gt}$ denotes the ground truth class. Similarly, $p^{\{x,y,w,h\}}$ stands for coordinate and box size predictions while $p^{\{x,y,w,h\}}_{gt}$ represents their ground truth targets. CE stands for the cross entropy loss \cite{girshick2015fast} to train box classifiers, and SmoothL1 represents the smooth L1 loss \cite{girshick2015fast} which is employed to train box regressors. Recently, focal loss \cite{lin2017focal} is often used to replace the cross entropy loss when the number of proposals is extremely large \cite{wang2020focalmix,zlocha2019improving}. Both one-stage and two-stage approaches require backbone networks to extract image features, where 2D and 3D deep neural networks are used according to the types of input data. Specifically, for 2D tasks, VGG-16 \cite{khosravan2018s4nd} and ResNet \cite{li2019mvp} are two widely adopted architectures. For 3D tasks, 3D ResNet \cite{wang2020focalmix,zlocha2019improving} and 3D U-Net \cite{liao2019evaluate} are two representatives. 

In this paper, we propose SSMD which incorporates  medical image detection with semi-supervised learning. Compared to semi-supervised classification/segmentation, SSMD focuses more on instance regions instead of the whole image in classification or individual pixels in segmentation. Accordingly, to better regularize instance regions in detection, our SSMD addresses the importance of adding consistency to instance locations which are usually ignored in semi-supervised classification/segmentation.}

In the following we describe three major contributions of SSMD: the adaptive consistency cost function, the noisy residual block and the instance-level adversarial perturbation strategy. We provide an overview in Fig.~\ref{framework} in which a student-teacher framework is employed to generate predictions for shared inputs with different perturbation strategies. 
For labeled images, the proposed method uses an adaptive consistency cost and the supervised loss. For unlabeled data, only the adaptive consistency cost is used. The consistency loss is calculated with the predicted proposals at each spatial position and each scale.

To make use of unlabeled images, it is necessary to mine the data to generate intrinsic supervision signals which can be further incorporated into the training process. Nowadays, most semi-supervised deep learning approaches~\cite{laine2016temporal,tarvainen2017mean,miyato2018virtual} focused on improving image classification results by keeping consistency within perturbed pairs. They require paired inputs where each pair contains the same image with different perturbation strategies. After feeding these pairs to neural networks, semi-supervised approaches force the outputs of each pair to be as close as possible. The most common perturbation methods can be summarized as: translation~\cite{laine2016temporal}, rotation~\cite{tarvainen2017mean} and horizontal flip~\cite{laine2016temporal,tarvainen2017mean,miyato2018virtual}. In this paper, we propose three more perturbation approaches: noisy residual block in feature space, instance-level adversarial perturbation and cutout in image space.

\subsection{Adaptive Consistency Cost}
\begin{figure}[!t]
	\centering
	\includegraphics[width=1.0\columnwidth]{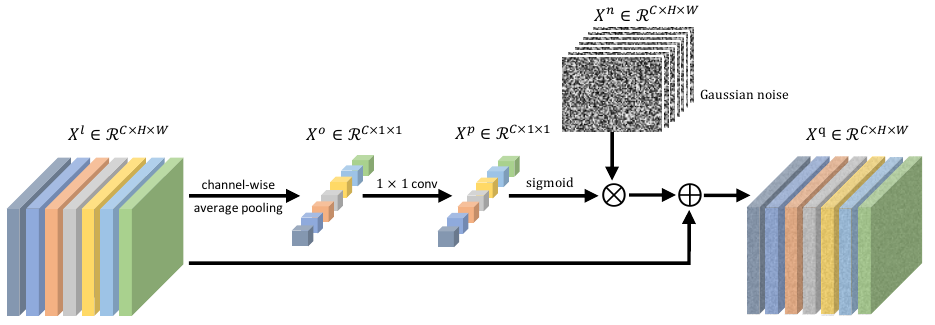}
	\caption{Illustration of the proposed noisy residual block. The proposed module adds noise perturbations to a residual block. Note that different colors mean different channels. $\otimes$ stands for channel-wise multiplication while $\oplus$ represents channel-wise addition.}
	\label{nc}
\end{figure}
As shown in Fig~\ref{framework}, the proposed SSMD model contains a student detector and a teacher detector where each network contains a feature pyramid network \cite{lin2017feature}. We adopt a parameter sharing approach proposed by \cite{tarvainen2017mean} where the teacher model uses the exponential moving average (EMA) weights of the student model. We denote the weights of the teacher model and the student model as $\theta_{t}$ and $\theta_{s}$ respectively. $\theta^{n}_{t}$ denotes the weights of the teacher network at training step $n$ and is updated as follows:
\begin{equation}
\begin{aligned}
\theta^n_t = \alpha\theta^{n-1}_{t} + (1-\alpha)\theta^{n}_{s}
\label{ema_formula}
\end{aligned}
\end{equation}
where both $\theta_0^t$ and $\theta_0^s$ are independently initialized. During the training stage, for the student branch we apply random rotation and then randomly mask out some rectangle regions, which is known as cutout \cite{devries2017improved}. As for the teacher branch, we first apply horizontal flip and cutout to the augmented input of the student branch, and then add instance-level adversarial perturbation to it. Different from CSD \cite{jeong2019consistency}, we propose to utilize an adaptive version of consistency cost to exploit unlabeled images and synthesize adversarial samples. The detector is based on RetinaNet which is to predict the positions of proposals relative to pre-defined anchors. $p^x, p^y, p^w, p^h$, which are outputs of the proposed detector, denote four scale factors:
\begin{equation}
\begin{aligned}
p^x &= (x - x^a) / w^a,\ p^y = (y - y^a) / h^a, \\
p^w &= \text{log}(w/w^a),\ p^h = \text{log}(h/h^a),
\end{aligned}
\end{equation}
where $x$, $y$ are the coordinate of a proposal center. $w$ and $h$ represent the width and the height of a proposal. $x^a$, $y^a$, $w^a$ and $h^a$ are variables for default anchors. Let $p^c$ denote the predicted probability distribution of different categories (after softmax). The whole procedure of the proposed semi-supervised medical detection is provided in Algorithm \ref{procedure}. We first apply different perturbations to a batch of labeled images $\mathcal{X}$ for two branches, respectively. After the forward pass, we obtain the predictions of classes and box coordinates. For each labeled image in $\mathcal{X}$, its supervised loss ($\text{loss}_{sup}$, shown in Line 8 of Algorithm \ref{procedure}), which consists of a cross entropy loss ($\text{CE}$) and a smooth L1 loss ($\text{SmoothL1}$), can be directly calculated between the prediction and the ground truth. 

To regularize the final predictions between the labeled images $\mathcal{X}$ and the unlabeled images $\mathcal{\tilde{X}}$, we apply a consistency cost which includes a KL-Divergence loss (KL) and a mean squared error (MSE) loss, as shown in Line \ref{loss_cont} of Algorithm~\ref{procedure}. { 
Assume that the output distributions of the teacher and the student models are close. Then KL loss is adopted for classification consisitency, to measure the output difference between the teacher and the student networks. 
For location consistency, we follow the setting of CSD\cite{jeong2019consistency} and adopt MSE loss.} 
Specifically, our proposed adaptive cost function contains a {dynamic instance weight} $\text{W}(p^c_s, p^c_t)$ which is defined as:
\begin{equation}
	\begin{aligned}
		\text{W}(p^c_s, p^c_t) = \frac{(1-p^c_s[\text{0}])^2+(1-p^c_t[\text{0}])^2}{2},
	\end{aligned}
\end{equation}
where $p^c_s[0]$ refers to the probability belonging to the background category, predicted by the student network. $p^c_t[0]$ shares the same definition and is predicted by the teacher network. {In our implementation, we treat the features of different levels equally in the adaptive consistency cost. For each feature level, the weight of the adaptive cost is equal to 1.} The MSE loss displayed in Line 11 and Line 12 of Algorithm \ref{procedure} is computed as:
\begin{equation}\label{MSE}
\begin{aligned}
\text{MSE}(p^{\{x,y,w,h\}}_s, p^{\{x,y,w,h\}}_t)=&\ \text{MSE}(p^x_s,p^x_t)+\text{MSE}(p^y_s,p^y_t)\\
&+\text{MSE}(p^w_s,p^w_t)+\text{MSE}(p^h_s,p^h_t),
\end{aligned}
\end{equation}
where $p_s$ and $p_t$ are the predictions of student model and teacher model, respectively. For the prediction of unlabeled data $\tilde{p}$, we calculate its MSE loss $\text{MSE} (\tilde{p}^{\{x,y,w,h\}}_s, \tilde{p}^{\{x,y,w,h\}}_t)$ in a similar way with Equation (\ref{MSE}).
Note that during the inference stage, only the student network $f_{\theta_s^N}(\cdot)$ is used to perform final predictions.

The proposed adaptive consistency cost takes into account the predicted confidence of proposals at each spatial position. Given a proposal with high foreground probability, it would result in a  higher weight of the consistency cost than those of easily recognized background samples. This mechanism helps the model apply more regularization effects to objects instead of the meaningless background. In practice, this adaptive cost is applicable to both labeled and unlabeled medical images, making proposed detector more effective in the setting of small amounts of labeled data.

\subsection{Noisy Residual Block}
In this part we propose \textit{noisy residual block} that adds noise to intermediate feature maps. The proposed noisy residual block can be regarded as a perturbation strategy working in a feature space. As shown in Fig.\ref{nc}, we modify the classical residual block used in \cite{he2016deep} and append an attention-based mechanism. We name the proposed module noisy residual block, since it introduces noise perturbations to a residual block. More details are in the following.

The input to layer $l$ is denoted as $X^l\in \mathcal{R}^{C\times H\times W}$. The proposed noisy residual block first applies a channel-wise average pooling to $X^l$ and then adopts a $1\times1$ convolutional operation:
\begin{equation}
	\begin{aligned}
		X^p = {conv}({AvgPool}(\it X^l)).
	\end{aligned}
\end{equation}
where $X^p \in \mathcal{R}^{C\times 1 \times 1}$ and \emph{AvgPool} is the abbreviation of global average pooling. For each layer $l$, we sample a Gaussian noise map $X^n \in \mathcal{R}^{C\times H\times W}$ where each component is drawn from a Gaussian distribution $\mathcal{N}(\mu, \sigma)$. $\mu$ and $\sigma$ stand for the mean and standard deviation, respectively. Meanwhile, we employ a scaled sigmoid function to normalize $X^p$. A channel-wise multiplication is performed between $X^p$ and $X^n$. Finally, $X^q$ can be computed by adding the multiplication result to the input feature $X^l$:

\begin{equation}
	\begin{aligned}
	X^q = (X^n \otimes {sigmoid}\ (\gamma \it X^p)) \oplus X^l.
	\end{aligned}
\end{equation}
where $\gamma$ is a scale factor. Here we employ a \emph{sigmoid} function to adaptively control the noise level of different channels in the noise perturbation. $X^q$ serves as the output of the noisy residual block and will be passed to following layers.

An intuitive understanding of the noisy residual block is to add ``appropriate'' noise to intermediate representations. For example, shallow layers are supposed to have wild noise as they are foundations of the whole network. We believe the degree of the embedded noise should be determined and can be learned by the representations themselves. Motivated by this idea, the noisy residual block learns channel-wise attentions to apply channel-dependent noise to feature maps. Moreover, we employ a residual connection to maintain the stability of the training process.

\subsection{{Instance-level Adversarial Perturbation based on Consistency Regularization}}
\label{secadv}
{Adversarial training has been widely adopted as a useful way to improve semi-supervised classification and segmentation. In contrast, the detection problem focuses more on instances instead of pixels in classification or segmentation. Thus, the methods designed for classification/segmentation may not be suitable for detection because they treat all pixels equally. In this section, we propose an instance-level adversarial perturbation strategy to address this issue.}

Let $r_{adv}$ denote the adversarial perturbations added to the input image. In each training iteration, $r_{adv}$ is first initialized from a normalized Gaussian distribution and has the same shape as $\mathcal{X}$ and $\mathcal{\tilde{X}}$. Then, a scaled $r_{adv}$ is added to the original image as:
\begin{equation}
\begin{aligned}
\text{Adv.}(\mathcal{X})=\mathcal{X}+\xi r_{adv},
\end{aligned}
\end{equation}
where $\xi$ is a scale factor satisfying $0< \xi \le 1$. Classical adversarial examples work by causing classifiers to predict a wrong category. However, in SSMD, the goal of adding adversarial perturbations is to increase the difficulty of performing consistency regularization. Note that similar computation process can also be applied to $\mathcal{\tilde{X}}$.

We follow \cite{goodfellow2014explaining} to synthesize a well perturbed input. We pass $\{\mathcal{X}, \mathcal{\tilde{X}}\}$ and $\{ \text{Adv.}(\mathcal{X}), \text{Adv.}(\mathcal{\tilde{X}})\}$ to student and teacher networks respectively, to obtain the consistency loss $\text{loss}_{cont}$ (shown in Line \ref{loss_cont} of Algorithm \ref{procedure}). {Only the high-confidence predictions are used to compute the consistency loss for gradient backward when applying adversarial perturbation. The gradient $g$ and the adversarial perturbations $r_{adv}$ are computed as:
\begin{equation}
\begin{aligned}
g &= \nabla_{r_{adv}}\text{loss}_{cont} * \mathds{1}[\textstyle\sum p^c_s\ ||\ \textstyle\sum \tilde{p}^c_s > \tau ],\\
r_{adv} &= \epsilon \frac{g}{||g||}.
\label{adv_conf}
\end{aligned}
\end{equation}
where the symbol $\textstyle\sum$ denotes the sum of all foreground classes.
$\mathds{1}[\cdot]$ is an indicator function which equals 1 when $\textstyle\sum p_s^c$ or $\textstyle\sum \tilde{p}_s^c$ is larger than a given threshold $\tau$. $\epsilon$ is the strength of perturbation, controlling the magnitude of $r_{adv}$. $||\cdot||$ stands for L2 normalization. After computing Equation (\ref{adv_conf}), $r_{adv}$ is added to $\mathcal{X}$ to obtain the final perturbed input. In general, it requires an additional forward and backward pass to synthesize the perturbed input image before we feed these final inputs to the detection network. Such process is to maximize the effect of $r_{adv}$ on $\text{loss}_{cont}$, and can be viewed as an adversarial process.}

Similar to the adaptive cost, we design instance-level perturbation to amplify the influences of high-confidence foreground proposals while reducing the impacts of low-confidence ones. In practice, foreground pixels receive heavy adversarial noise while the perturbation of background pixels has much smaller magnitude. Such implementation makes the consistency loss focus more on foreground objects, producing effectively perturbed inputs.

\section{Experiments}
In this section, we first conduct ablation studies to better understand the strengths of different modules in the proposed method SSMD. Moreover, we design comprehensive experiments to verify the effectiveness of SSMD on various settings.

\subsection{Dataset}
The experiments are conducted on a nuclei dataset and a lesion database.
For both datasets, we manually and randomly split the training set into labeled data and unlabeled data with fixed ratios in order to fit the setting of semi-supervised learning. 

\textbf{Nuclei Dataset }
In our experiments, we adopt the nuclei dataset introduced by 2018 Data Science Bowl\footnote{\url{https://www.kaggle.com/c/data-science-bowl-2018}} (DSB, hosted by Kaggle). The dataset was acquired under a variety of conditions and includes nuclei images of different cell types, magnifications, and imaging modalities. The training set contains 522 nuclei images (80\%) while the validation set has about 60 images (10\%). The rest images are used for testing. On average, each image contains about 45 cells which are enough to train a robust nuclei detector. In practice, we only assign labels to some training images and take the other training images as unlabeled data. The evaluation metric is mAP

\textbf{DeepLesion Dataset }
We also present experimental results on DeepLesion \cite{yan2018deeplesion} which is a large-scale public dataset containing 32,120 axial Computed Tomography (CT) slices of 10,594 studies collected from 4,427 patients. The dataset has 32,735 annotated lesion instances in total. Each slice contains 1$\sim$3 lesions. 
The additional slices above and beneath a target slice are regarded as relevant contexts of the target slice. These additional slices are of 30 mm. In most cases, a slice is 1 or 5 mm thick. The dataset covers a wide scope of lesions from lung, liver, mediastinum (essentially lymph hubs), kidney, pelvis, bone, midsection and delicate tissue. 
Following \cite{zlocha2019improving,li2019mvp}, we test our proposed method on official testing set (15\%) and report the sensitivity at 4 false positives (FPs). We directly use the training and validation set officially provided by DeepLesion.

\subsection{Implementation Details}
For DSB dataset, the proposed detector is built on top of an ImageNet-pretrained ResNet-50 which has five scales. Nine default anchors are adopted in each scale. The size of input images is 448$\times$448. The batch size is 8. All models are trained for 100 epochs. Adam is utilized as the default optimizer with 1e-5 as the initial learning rate, which is then divided by 10 at the 75th ($100\times \frac{3}{4}$) epochs. For the supervised baseline, image rotation and horizontal image flipping are considered as default augmentation strategies. It is worth noting that the hyperparameter $\lambda$ of consistency loss (shown in Line~\ref{lambda} of Algorithm \ref{procedure}) plays an important role during the training stage. We first gradually increase the value of $\lambda$ to 1 in the first quarter of the training, and then decrease it to 0 in the last quarter. 
The formal definition of $\lambda$ is:
\begin{equation}
\lambda = \left\{
\begin{array}{lr}
e^{-5(1-\frac{4j}{N})^2}, & 0 \leq j < \frac{N}{4} \\
1, & \frac{N}{4} \leq j < \frac{3N}{4}\\
e^{-12.5(1-\frac{7(N-j)}{N})^2}, &\frac{3N}{4} \leq j \leq N
\end{array}
\right.
\end{equation}
where $N$ is the number of training iterations and $j$ is the iteration index. Similarly, for DeepLesion dataset we simply follow the preprocessing method in \cite{zlocha2019improving} to resize each slice into 512$\times$512 pixels whose mean voxel-spacing is 0.802mm. We first clip the Hounsfield units (HU) to [-1100, 1100] and then normalize them to [-1,1]. We compute the mean and standard deviation of the whole training set and use them to further normalize input slices. For both datasets, we set $\gamma$ to 0.9 and the degree of random rotation is set to 10 degrees.

\begin{table*}[!t]
    \tiny
	\centering
	\subfloat[][Backbone: ResNet-50]{{
	\begin{tabular}{C{1.5cm}C{3cm}L{1cm}L{1cm}L{1cm}L{1cm}L{1cm}}
	\hline
	Dataset & Method & 10\% & 20\% & 30\% & 40\% & 50\% \\
	\hline
	\multirow{5}{*}{DSB} & Supervised & 39.3 & 47.3 & 53.1 & 58.9 & 62.3\\
	& Pseudo-labeling & 46.8 & 53.5 & 57.2 & 62.5 & 65.5\\
	& CSD & 45.7 & 54.0 & 58.5 & 62.9 & 65.2\\
	& FocalMix & 48.9 & 56.2 & 60.1 & 64.0 & 67.0\\
	& SSMD & \textbf{52.2}$\scriptstyle{+3.3}$ & \textbf{59.1}$\scriptstyle{+2.9}$ & \textbf{62.3}$\scriptstyle{+2.2}$ & \textbf{65.9}$\scriptstyle{+1.9}$ & \textbf{68.8}$\scriptstyle{+1.8}$\\
	\hline
	DSB & Fully-supervised & \multicolumn{5}{c}{76.7} \\
	\hline
	- & $p$-value & 0.0154 & 0.0232 & 0.0195 & 0.0274 & 0.0242\\
	\toprule
	\multirow{5}{*}{DeepLesion} & Supervised & 48.0 & 54.9 & 61.6 & 67.3 & 71.5\\
	& Pseudo-labeling & 55.6 & 59.8 & 64.7 & 70.2 & 73.6\\
	& CSD & 55.3 & 59.5 & 65.2 & 71.0 & 74.0\\
	& FocalMix & 57.6 & 62.2 & 66.9 & 72.4 & 75.3\\
	& SSMD & \textbf{60.4}$\scriptstyle{+2.8}$ & \textbf{64.1}$\scriptstyle{+1.9}$ & \textbf{68.4}$\scriptstyle{+1.5}$ & \textbf{74.1}$\scriptstyle{+1.7}$ & \textbf{76.5}$\scriptstyle{+1.2}$\\	
	\hline
	DeepLesion& Fully-supervised & \multicolumn{5}{c}{86.5} \\
	\hline
	- & $p$-value & 0.0096 & 0.0104 & 0.0083 & 0.0074 & 0.0098\\
	\hline
	
	\end{tabular}
	}\label{<figure1>}}\\
	\subfloat[][Backbone: ResNet-101]{{
	\begin{tabular}{C{1.5cm}C{3cm}L{1cm}L{1cm}L{1cm}L{1cm}L{1cm}}
	\hline
	Dataset & Method & 10\% & 20\% & 30\% & 40\% & 50\% \\
	\hline
	\multirow{5}{*}{DSB} & Supervised & 42.1 & 50.6 & 55.7 & 61.0 & 64.5\\
	& Pseudo-labeling & 49.0 & 55.6 & 59.4 & 64.7 & 66.8\\
	& CSD & 48.2 & 55.8 & 59.9 & 65.1 & 66.2\\
	& FocalMix & 51.0 & 58.4 & 62.3 & 65.9 & 67.4\\
	& SSMD & \textbf{54.8}$\scriptstyle{+3.8}$ & \textbf{61.7}$\scriptstyle{+3.3}$ & \textbf{64.9}$\scriptstyle{+2.6}$ & \textbf{68.2}$\scriptstyle{+2.3}$ & \textbf{69.4}$\scriptstyle{+2.0}$\\	
	\hline
	DSB & Fully-supervised & \multicolumn{5}{c}{78.3} \\
	\hline
	- & $p$-value & 0.0121 & 0.0135 & 0.0093 & 0.0101 & 0.0089\\
	\bottomrule
	\multirow{5}{*}{DeepLesion} & Supervised & 50.8 & 57.3 & 64.0 & 69.5 & 73.0\\
	& Pseudo-labeling & 57.7 & 61.5 & 66.6 & 72.1 & 74.7\\
	& CSD & 57.1 & 61.8 & 67.4 & 73.2 & 75.1\\
	& FocalMix & 59.8 & 64.3 & 69.0 & 74.5 & 76.5\\
	& SSMD & \textbf{62.5}$\scriptstyle{+2.7}$ & \textbf{66.4}$\scriptstyle{+2.1}$ & \textbf{70.7}$\scriptstyle{+1.7}$ & \textbf{76.1}$\scriptstyle{+1.6}$ & \textbf{77.9}$\scriptstyle{+1.4}$\\
	\hline
	DeepLesion & Fully-supervised & \multicolumn{5}{c}{87.9}\\
	\hline
	- & $p$-value & 0.0073 & 0.0082 & 0.0065 & 0.0061 & 0.0078\\
	\hline
	
	\end{tabular}
	}\label{<figure2>}}

	\caption{Comparison with baseline approaches under different labeled ratios. For all SSMD experiments, we apply cutout to their inputs. We report experimental results on both ResNet-50 and ResNet-101. All fully-supervised baselines make use of 100\% labeled data. The input resolution of DSB is set to 448$\times$448 while the input size of DeepLesion is 512$\times$512. The best results are in bold and we also display the relative improvements compared to the second best results. All $p$-values are computed between SSMD and FocalMix.}
	\label{baseline}
\end{table*}

\subsection{Baselines}
\textbf{Supervised and fully-supervised detectors } For supervised detector, we only use the labeled data to train deep models. As for the fully-supervised baseline, we train our nuclei and lesion detector using the whole training set. In practice, for both two approaches, we save checkpoints based on their performance on validation sets. These models are then used to perform the test whose results are reported in the following.

\textbf{Pseudo-labeling } An intuitive thought in semi-supervised classification for utilizing unlabeled data is to use a trained model to make predictions, which can also be applied to object detection. However, considering the fact that existing deep learning based detectors usually produce unreliable results given a small amount of training data, we have to cherry-pick optimal predictions from dozens of region candidates. It is a laborious and tedious process. Therefore the strategy used in \cite{sohn2020simple} is applied to filter low-confidence candidates with a high confidence threshold. Pseudo labels with their corresponding images and labeled data are utilized to train a new detector from scratch.

\textbf{Consistency-based Semi-Supervised learning method for object Detection (CSD) } CSD \cite{jeong2019consistency} employs consistency constraints as a tool to improve detection performance by making full use of unlabeled data. For each input image, CSD first applies horizontal flip to construct an input pair which is fed to a siamese network to obtain two sets of predictions. Constraints are then added to regularize these predictions and serve as an additional supervision for unlabeled images. Compared to CSD, our proposed method emphasizes the robustness of predictions under various perturbations.

\textbf{FocalMix} FocalMix \cite{wang2020focalmix} was the the first approach to investigate the problem of semi-supervised learning for medical image detection. FocalMix is based on MixMatch \cite{berthelot2019mixmatch} whose idea is similar to  pseudo-labeling. Mixup \cite{zhang2017mixup} is used as the main augmentation strategy in labeled training set and a soft-target focal loss is proposed to leverage soft targets. The original version of FocalMix is a 3D detector while we extends it to a 2D version.

{\textbf{Implementation of baselines} For fairness, all baselines share the same detection backbone (i.e., RetinaNet), input sizes and training strategies with those of our SSMD, if not specified otherwise. For Pseudo-labeling, we set the confidence threshold to 0.9 following the instruction from \cite{sohn2020simple} and use one-shot encoding to represent pseudo-labels. For CSD, we use KL divergence and MSE loss to regularize detector's predictions. 
Since the code of FocalMix has not been released, we directly implement a 2D detector (based on RetinaNet) following the official paper\cite{wang2020focalmix}.
That is to say, we employ the same mixup strategies including image-level mixup and object-level mixup to augment training data. For mixup hyperparameters, we set $\eta$ to 0.2. We also implement the soft-target focal loss together with the sharpening operator, where $\alpha_0$ is set to 0.05, $\alpha_1$ is set to 0.95, $\gamma$ is set to 2.0 and the sharpen temperature factor $T$ is set to 0.7. We have tried to train FocalMix for \{100,200,400,800\} epochs, where we find that training for 200 epochs achieves the best performance in both DSB and DeepLesion.}

\subsection{Comparison with the Baselines}
The experimental results of supervised and semi-supervised approaches are reported in Table \ref{baseline}. We conduct experiments on both DSB and DeepLesion datasets with different network architectures and different labeled ratios. 

The first observation is that the proposed SSMD can consistently outperform other semi-supervised baselines by a significant margin in different datasets and labeled ratios. Particularly, SSMD has greater advantage in small labeled ratios, such as 0.1 \& 0.2. For example, SSMD outperforms FocalMix by more than 3 points in DSB when the labeled ratio is 10\% while such gap becomes smaller as the labeled ratio increases. Similar phenomena can also be observed in DeepLesion dataset. We argue that the significance in small labeled ratios may be due to the fact that more perturbations lead to stronger regularization, preventing deep neural networks from the overfitting scenario which is pretty common given a small number of training data. Such argument is supported by the following observation: the performance gap between SSMD and other semi-supervised methods is larger in DSB than the gap in DeepLesion. Because DeepLesion is a large-scale dataset and even 10\% of it has thousands of training samples which definitely help deep learning models avoid overfitting.

Another observation is that FocalMix consistently surpasses pseudo-labeling and CSD while pseudo-labeling performs as well as CSD in most cases. Typically, pseudo-labeling seems to obtain better results in small labeled ratios whereas CSD performs better in large ratios. Such trend is obvious in DeepLesion but becomes less visible in DSB. Nonetheless, all semi-supervised methods significantly outperform the supervised baseline, which verifies that semi-supervised methods are considerably helpful with small ratios of labeled data.

\begin{figure}[!htp]
	\centering
	\subfloat[DSB (ResNet-50)]{\includegraphics[width=0.48\columnwidth]{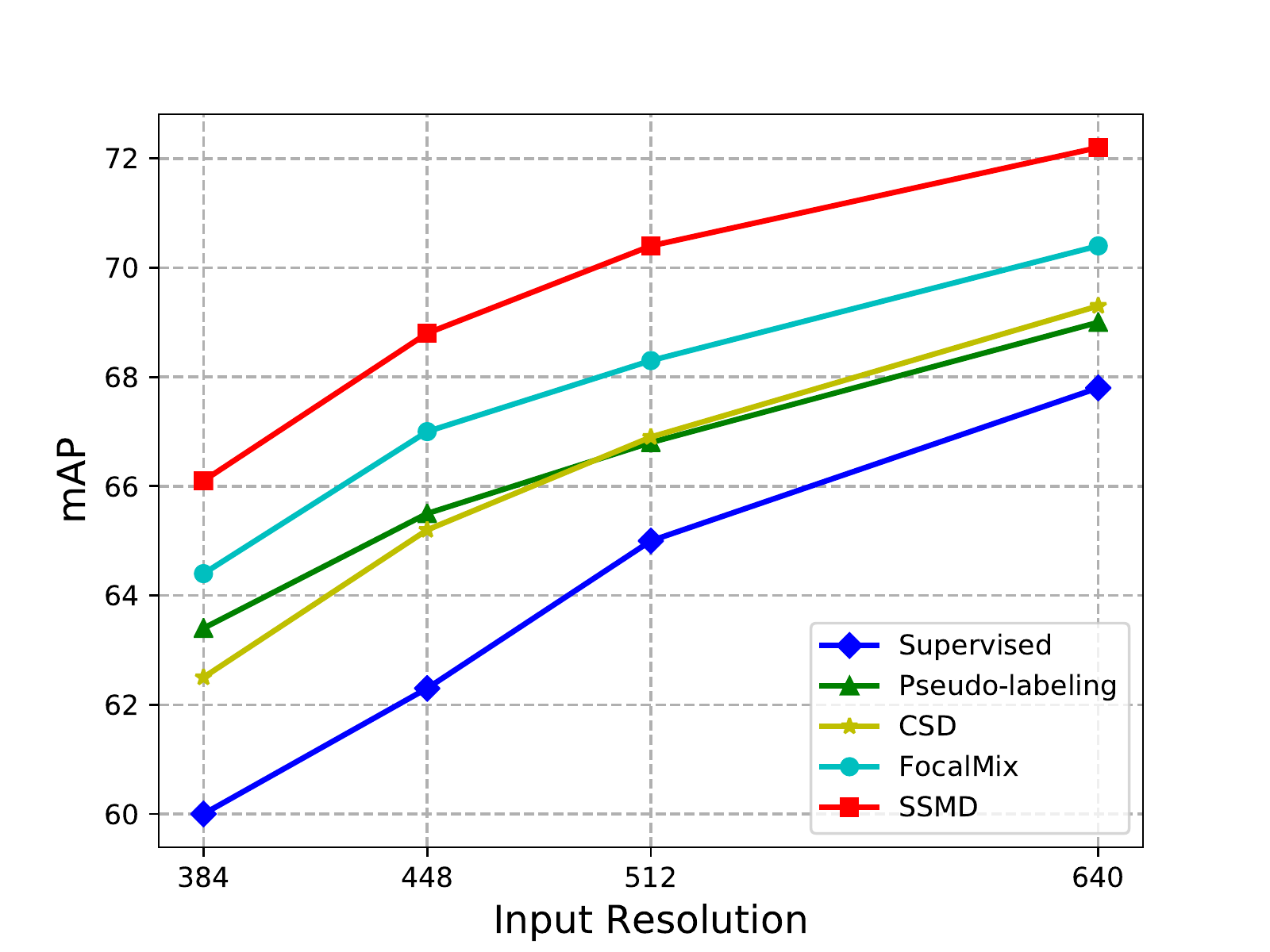}}
	\ \ \ 
	\subfloat[DSB (ResNet-101)]{\includegraphics[width=0.48\columnwidth]{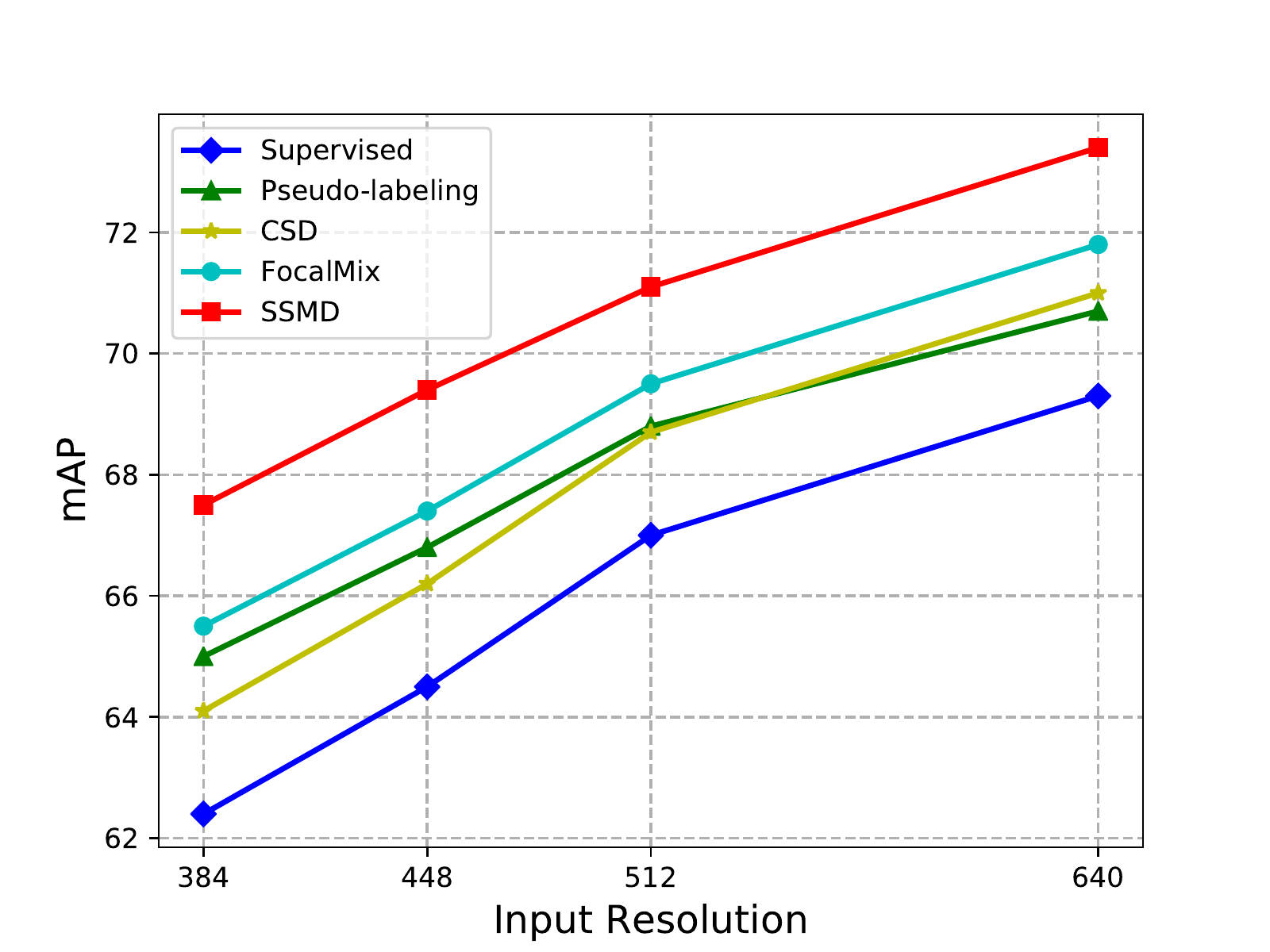}}
	\ \ \ 
	\subfloat[DeepLesion (ResNet-50)]{\includegraphics[width=0.48\columnwidth]{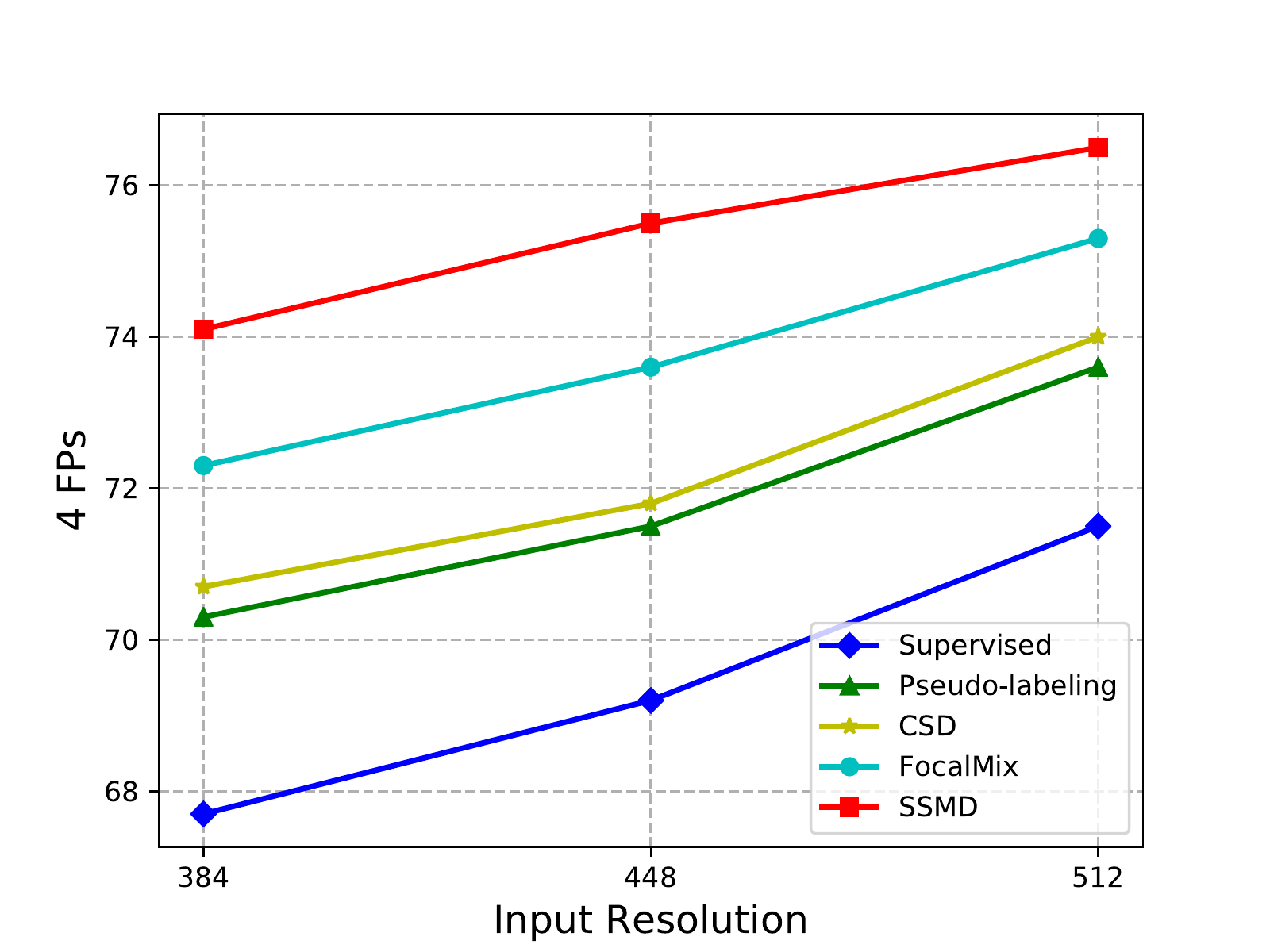}}
	\ \ \ 
	\subfloat[DeepLesion (ResNet-101)]{\includegraphics[width=0.48\columnwidth]{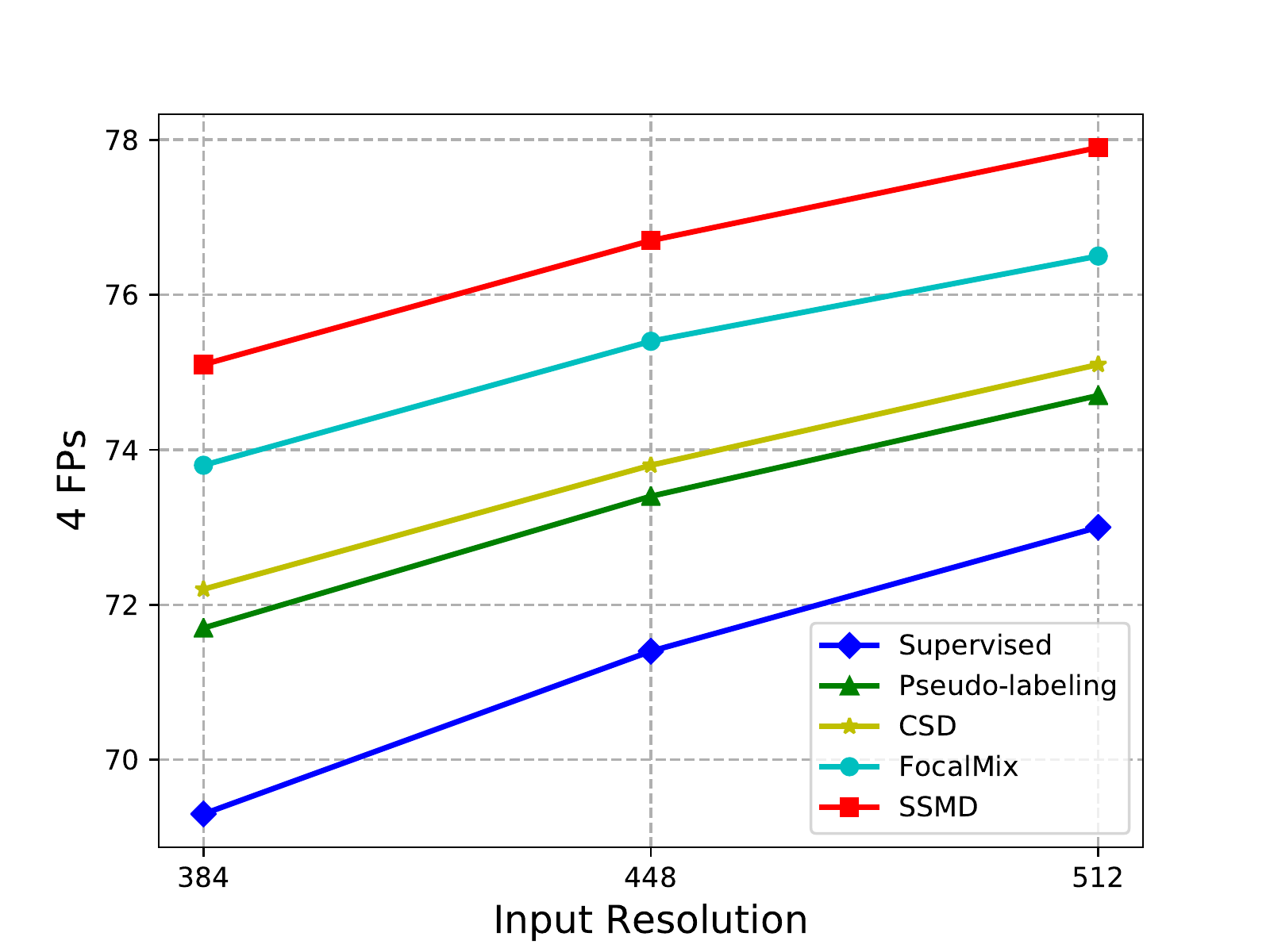}}
	\caption{Comparison of different semi-supervised detectors with various input resolutions. mAP and 4 FPs are used for DSB and DeepLesion as evaluation metrics. The labeled ratio is 50\%.}
	\label{resolution}
\end{figure}

In Fig.\ref{resolution}, we conduct experiments on different approaches with various input resolutions and two different network architectures. We can see that SSMD outperforms FocalMix significantly in all resolutions on both DSB and DeepLesion datasets. If we compare CSD with pseudo-labeling, we can see that although pseudo-labeling can outperform CSD on DSB dataset under small labeled ratios, it cannot beat CSD on DeepLesion dataset. For SSL with different network architectures, we can find that the gap between SSMD and other baselines becomes smaller when network becomes deeper (ResNet-50 vs. ResNet-101). We argue that deeper models may reduce the relative improvements of most of SSL algorithms on both DSB and DeepLesion datasets.

\begin{table}
	\centering
	\begin{tabular}{ccc|cc}
	ACC & NRB & IAP & DSB & DeepLesion\\
	\hline
	\hline
	& & & 65.2 & 74.0 \\
	\hline
	\checkmark & & & 66.7 & 75.1 \\
	\checkmark & \checkmark & & 68.0 & 75.9 \\
	\checkmark & \checkmark & \checkmark & 68.8 & 76.5\\
	\hline
	\multicolumn{3}{c|}{$p$-value} & 0.0227 & 0.0328\\
	\end{tabular}
	\caption{An incremental study of all proposed modules. The default input resolutions for DSB and DeepLesion are $448\times 448$ and $512\times 512$, respectively. \textbf{ACC} stands for adaptive consistency cost, \textbf{NRB} denotes the proposed noisy residual block and \textbf{IAP} represents the instance-level adversarial perturbation. The p-values are calculated between the top-2 models in each column.}
	\label{incremental}
\end{table}

\begin{table}[!t]
	\centering
	\begin{tabular}{c|c|cccc}
		Dataset & ACC & 384 & 448 & 512 & 640\\
		\hline
		\hline
		DSB & & 62.5 & 65.2 & 66.9 & 69.3 \\ 
		DSB & \checkmark & 63.8 & 66.7 & 68.6 & 71.1 \\
		\hline
		\multicolumn{2}{c|}{$p$-value} & 0.0174 & 0.0133 & 0.0121 & 0.0107\\
		\hline
		DeepLesion & & 70.7 & 71.8 & 74.0 & -\\
		DeepLesion & \checkmark & 71.6 & 72.8 & 75.1 & - \\
		\hline
		\multicolumn{2}{c|}{$p$-value} & 0.0154 & 0.0143 & 0.0127 & -\\
	\end{tabular}
	\caption{Influence of the proposed adaptive consistency cost (ACC). Note that for SSMD without the adaptive consistency cost, we directly employ the consistency loss used in CSD.}
	\label{acc}
\end{table}

\begin{table*}[!t]
	\scriptsize
	\centering
	\begin{tabular}{cccc|cccc|ccc}
		\multirow{2}{*}{Dropout} & \multirow{2}{*}{Dropblock} & \multirow{2}{*}{Noise} & \multirow{2}{*}{NRB} & \multicolumn{4}{c|}{DSB (mAP)} & \multicolumn{3}{c}{DeepLesion (4 FPs)}\\
		& & & & 384 & 448 & 512 & 640 & 386 & 448 & 512\\
		\hline
		\hline
		& & & & 62.5 & 65.2 & 66.9 & 69.3 & 70.7 & 71.8 & 74.0\\
		\checkmark & & & & 62.8 & 65.0 & 66.3 & 69.2 & 70.1 & 71.3 & 73.6 \\ 
		& \checkmark & & & 62.9 & 65.6 & 67.3 & 69.4 & 71.0 & \underline{72.3} & 74.2\\ 
		& & \checkmark& & 62.5 & 65.3 & 66.9 & 69.1 & 70.8 & 71.7 & 73.9 \\ 
		& & & \checkmark & \textbf{63.6} & \textbf{66.2} & \textbf{67.8} & \textbf{70.1} & \textbf{71.9} & \textbf{72.8} & \textbf{74.8}\\	
		\hline
		\checkmark & & & \checkmark & 63.0 & \underline{65.8} & \underline{67.4} & \underline{69.5} & 70.8 & 72.1 & \underline{74.6} \\ 	
		& & \checkmark & \checkmark & \underline{63.2} & 65.0 & 66.9 & 69.1 & 71.2 & 71.3 & 74.3 \\ 
		\hline
		\multicolumn{4}{c|}{$p$-value} & 0.0324 & 0.0297 & 0.0331 & 0.0196 & 0.0203 & 0.0297 & 0.0406\\
	\end{tabular}
	\caption{Investigation of different perturbation strategies in feature space. We report experimental results on DSB and DeepLesion using different perturbation strategies in feature space with various input resolutions. The first line corresponds to the results of CSD. The best results are in bold while the second best are underlined. \textbf{NRB} stands for the noisy residual block. Here we do not use adaptive consistency cost, cutout or adversarial perturbation. All p-values are calculated between the top-2 models in each column.}
	\label{ab_nrb}
\end{table*}

\subsection{Ablation Study}
In this section, we first make an incremental study on all proposed modules. Then, we study how different settings and factors affect the proposed SSMD. These factors include the adaptive consistency cost (ACC), the perturbation strategies in feature space, the hyperparameters of adding adversarial perturbations ($\xi$, $\epsilon$ and $\tau$), the EMA factor $\alpha$, different consistency constraints ($c$, $x$, $y$, $w$ and $h$), different number and size of masks for applying cutout, and adding consistency loss to different numbers of feature scales. Considering that these hyperparameters could have different affects with different resolutions, we present the results with different input sizes. For all experiments, \emph{the default labeled ratio is 0.5 and the network architecture is ResNet-50}. We repeat each experiment for three times and report their average results. 

\subsubsection{{An Incremental Study of Proposed Modules}}
{In this part, we mainly conduct an incremental study of all proposed modules. Generally speaking, we can see that the proposed modules can consistently boost the overall performance in both DSB and DeepLesion datasets. More specifically, we can see that the adaptive consistency cost seems to bring larger improvements when compared to the noisy residual block and the instance-level perturbation, which are 1.5 points in DSB and 1.1 points in DeepLesion. After adding the noisy residual block, our SSMD can already outperform FocalMix by more than 0.5 point. Nonetheless, the instance-level adversarial perturbation finally helps SSMD to surpass FocalMix by a substantial margin in both DSB and DeepLesion datasets.}

\subsubsection{Investigation of Adaptive Consistency Cost}
We study the influence of adding adaptive consistency cost in Table \ref{acc}. It is obvious that the proposed adaptive consistency cost helps to boost the detection performance in all input resolutions. Specifically, the adaptive cost brings more than 1 point improvements on both DSB and DeepLesion datasets. Interestingly, we can see that as the input resolution increases, the effectiveness of the adaptive cost becomes more significant. For example, on DSB dataset when the input resolution is 640$\times$640, the relative improvements are 1.8 points which are larger than those of small input resolutions. Similar phenomena also appears on DeepLesion dataset. The underlying reason can be that bigger inputs usually bring more region proposals where our adaptive mechanism is more effective in amplifying the significance of high-confidence proposals.

\subsubsection{Investigation of Perturbation Strategies in Feature Space}
\begin{table*}[!t]
	\centering
	\subfloat[DSB]{
		\begin{tabular}{cc|C{1cm}C{1cm}C{1cm}C{1cm}}
			\multirow{2}{*}{$\xi$} & \multirow{2}{*}{$\epsilon$} & \multicolumn{4}{c}{Image Size}\\
			& & 384 & 448 & 512 & 640\\
			\hline\hline
			0 & 0 & 62.5 & 65.2 & 66.9 & 69.3\\
			1e-8 & 1.0 & 63.0 & 65.6 & 67.3 & 69.7 \\
			1e-7 & 1.0 & 63.2 & 65.8 & 67.6 & 70.2 \\
			5e-7 & 1.0 & \underline{63.5} & \underline{66.0} & \underline{67.8} & \underline{70.4} \\
			5e-7 & 2.0 & \textbf{63.7} & \textbf{66.2} & \textbf{68.2} & \textbf{70.5} \\
			1e-6 & 2.0 & 63.4 & 65.9 & 67.4 & 70.1 \\
			5e-7 & 3.0 & 63.3 & 65.8 & 67.2 & 70.3 \\
			\hline
			\multicolumn{2}{c|}{$p$-value} & 0.0743 & 0.0521 & 0.0634 & 0.0586\\
		\end{tabular}
		\label{adv(a)}
	}
	\quad\quad
	\subfloat[DeepLesion]{
		\begin{tabular}{cc|C{1cm}C{1cm}C{1cm}}
			\multirow{2}{*}{$\xi$} & \multirow{2}{*}{$\epsilon$} & \multicolumn{3}{c}{Image Size}\\
			& & 384 & 448 & 512 \\
			\hline\hline
			0 & 0 & 70.7 & 71.8 & 74.0 \\
			1e-7 & 1.0 & 71.1 & 72.2 & 74.3 \\
			1e-7 & 2.0 & 71.0 & 72.2 & 74.1 \\
			5e-7 & 1.0 & 71.3 & 72.5 & 74.4 \\
			5e-7 & 2.0 & \underline{71.5} & 72.9 & \underline{74.7} \\
			1e-7 & 2.0 & \textbf{71.7} & \textbf{73.2} & \textbf{74.9} \\
			1e-5 & 1.0 & 71.4 & \underline{73.0} & 74.5 \\
			\hline
			\multicolumn{2}{c|}{$p$-value} & 0.0627 & 0.0689 & 0.0754 \\
		\end{tabular}
		\label{adv(b)}
	}
	\caption{Ablation study of $\xi$ and $\epsilon$ in adversarial perturbation. When $\xi$ or $\epsilon$ is 0, our method is equivalent to CSD~\cite{jeong2019consistency}. We set $\tau$ to 0.95. Note that we do not use adaptive cost function, noisy residual block or cutout in these experiments. The best results are in bold while the second best are underlined. All p-values are calculated between the top-2 models in each column.}
	\label{adv}
\end{table*}
In Table \ref{ab_nrb}, we compare the performance of different perturbation strategies in feature space. Note that we did not use cutout, adaptive cost function or adversarial perturbation, so the results in Table \ref{ab_nrb} are lower than those in Table \ref{baseline}. The results of CSD is displayed in the first line of Table \ref{ab_nrb}.

We first adopt the most widely used strategy Dropout \cite{srivastava2014dropout} to regularize the feature space. Unfortunately, it seems that applying layer-wise dropout is not a good idea as it slightly degrades the performance when compared to the original CSD. We then give a trial to Dropblock \cite{ghiasi2018dropblock} which randomly masks rectangle regions in feature maps and was initially designed for object detection with natural images. Different from Dropout, Dropblock needs to be located at specific layers within the convolutional neural network. Following the instructions proposed by \cite{ghiasi2018dropblock}, we place Dropblock in conv4 and conv5 only. The results in Table \ref{ab_nrb} suggest that Dropblock achieves better performances than Dropout. Moreover, we try adding Gaussian noise (denoted as `Noise') to each layer, which performs slightly better than Dropout but still worse than Dropblock.

The proposed layer-wise noisy residual block (denoted as `NRB' in Table \ref{ab_nrb}) outperforms Dropblock remarkably regardless of datasets and input resolutions. The reason behind may be that the strength of added noise in the noisy residual block is learnable. It means that different feature channels have different levels of noise. This principle enables the neural network to learn the optimal manner to implement layer-wise noise. In addition, we try to integrate Dropout or Gaussian noise with the proposed noisy residual block. However, we found that these naive combinations slightly degrade the detection performance.

\subsubsection{Instance-level Adversarial Perturbation in Image Space}
As described in Section \ref{secadv}, the proposed adversarial perturbation has three hyperparameters: the scale factor $\xi$, the confidence threshold $\tau$ and the magnitude controlling factor $\epsilon$. In the following, we conduct comprehensive experiments to demonstrate the effect of each hyperparameter.

\textbf{Scale Factor $\xi$ and Confidence Threshold $\epsilon$ } Table \ref{adv} displays how we determine the exact values of $\xi$ and $\epsilon$. In Table \ref{adv} the proposed method SSMD considerably surpasses the baseline CSD with a naive perturbation implementation. In practice, we determine the values of $\xi$ and $\epsilon$ considering their influences on two datasets. For DSB dataset, the input resolution is set to $448\times 448$ while the input size of DeepLesion is $512\times 512$. For other input sizes, we simply use the values of $\xi$ and $\epsilon$ chosen in ablation study. As Table \ref{adv} displays, the setting of $\{$$\xi$=5e-7, $\epsilon$=2.0$\}$ achieves nearly the best performance in both datasets. Although the setting of $\{$$\xi$=1e-7, $\epsilon$=2.0$\}$ performs the best in DeepLesion dataset, we still adopt $\{$$\xi$=5e-7, $\epsilon$=2.0$\}$ as our default setting.

\begin{figure}[!t]
	\centering
	\includegraphics[width=0.8\columnwidth]{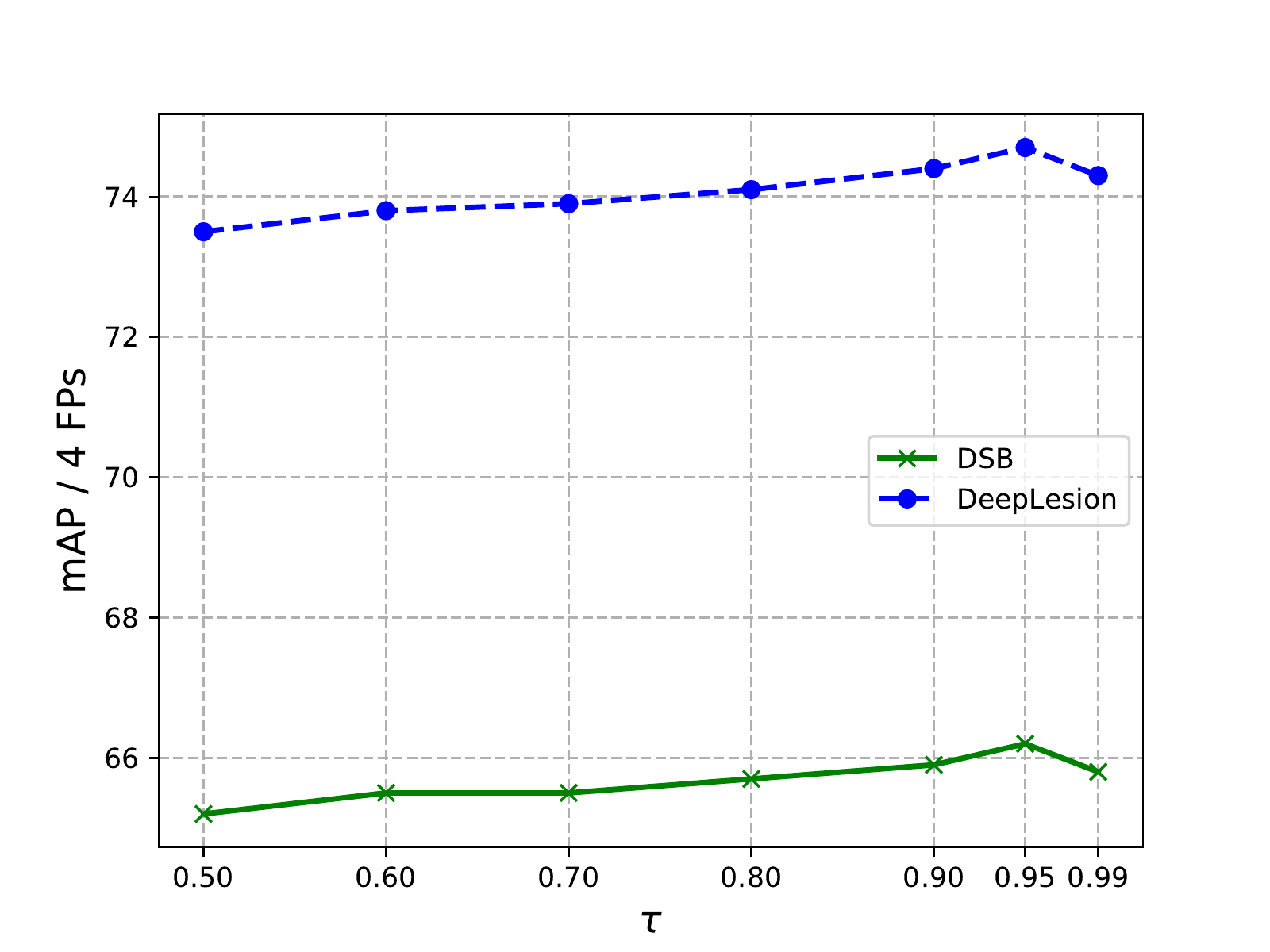}
	\caption{Investigation of different confidence thresholds. Here we do not apply cutout to network inputs. $\xi$ is set to 5e-7 while $\epsilon$ is set to 2.0. The default input resolutions for DSB and DeepLesion are $448\times 448$ and $512\times 512$, respectively. The $p$-values between top-2 choices in DSB and DeepLesion are 0.0437 and 0.0348, respectively.} 
	\label{tau}
\end{figure}
\begin{table}[!htp]
	\centering
	\begin{tabular}{c|ccccc|c}
		$\alpha$ & 0.8 & 0.9 & 0.95 & 0.99 & 0.999 & $p$-value\\
		\hline
		\hline
		DSB & 64.7 & 65.4 & 65.7 & 65.8 & 65.6 & 0.0347\\
		DeepLesion & 73.4 & 74.3 & 74.6 & 74.8 & 74.7 & 0.0286\\ 
	\end{tabular}
	\caption{Investigation of EMA factor. The default input resolutions for DSB and DeepLesion are $448\times 448$ and $512\times 512$, respectively. Note that we do not apply cutout or adversarial perturbation to network inputs and we also do not use adaptive cost function. All p-values are calculated between the top-2 models in each row.}
	\label{ema}
\end{table}
\begin{table*}[htp]
	\centering
	\begin{tabular}{ccccc|cccc|ccc}
		\multirow{2}{*}{$c$ }& \multirow{2}{*}{$x$} & \multirow{2}{*}{$y$} & \multirow{2}{*}{$w$} & \multirow{2}{*}{$h$} & \multicolumn{4}{c|}{DSB (mAP)} & \multicolumn{3}{c}{DeepLesion (4 FPs)} \\
		& & & & & 384 & 448 & 512 & 640 & 384 & 448 & 512 \\ 
		\hline\hline
		& & & & & 60.0 & 62.3 & 65.0 & 67.8 & 67.7 & 69.2 & 71.5\\
		\checkmark & & & & & 62.2 & 64.3 & 66.6 & 69.2 & 69.1 & 70.8 & 73.0 \\
		\checkmark & \checkmark & & & & 62.3 & 64.8 & 66.8 & 69.5 & 69.4 & 71.1 & 73.2\\
		\checkmark & \checkmark & \checkmark & & & 62.9 & 65.6 & 67.2 & 69.9 & 70.6 & 72.0 & 74.3\\
		\checkmark & \checkmark & \checkmark & \checkmark & & 62.9 & 65.7 & 67.4 & 69.9 & 71.0 & 72.4 & 74.5\\
		\checkmark & \checkmark & \checkmark & \checkmark & \checkmark & 63.3 & 66.0 & 67.6 & 70.2 & 71.5 & 72.6 & 74.7\\
		\hline
		\checkmark & \checkmark & \checkmark & \checkmark & \checkmark & 62.5* & 65.2* & 66.9* & 69.3* & 70.7* & 71.8* & 74.0* \\
		\hline
		\multicolumn{5}{c|}{$p$-value} & 0.0329 & 0.0272 & 0.0292 & 0.0314 & 0.0256 & 0.0236 & 0.0228\\
	\end{tabular}
	\caption{Investigation of adding consistency loss with different combinations of predictions ($c$ for confidence score, $h$ for height, $w$ for width, $x$ and $y$ for centering coordinate). * denotes the results of CSD. Note that we do not use cutout, noisy residual block and adversarial perturbation in these experiments. All p-values are calculated between the top-2 models in each column.}
	\label{loss}
\end{table*}

\begin{table*}[!t]
	\centering
	\subfloat[DSB]{
		\begin{tabular}{cc|C{1cm}C{1cm}C{1cm}C{1cm}}
			\multirow{2}{*}{$n$} & \multirow{2}{*}{$s$} & \multicolumn{4}{c}{Image Size}\\
			& & 384 & 448 & 512 & 640\\
			\hline\hline
			0 & 0 & 62.5 & 65.2 & 66.9 & 69.3 \\
			3 & 50 & 63.0 & 65.5 & 67.2 & 69.4\\
			3 & 70 & 63.0 & 65.4 & 67.0 & 69.2\\
			5 & 50 & \textbf{63.5} & 65.7 & 67.4 & 69.6\\
			5 & 70 & \textbf{63.5} & \textbf{66.2} & \underline{67.7} & \underline{69.8}\\
			7 & 50 & \underline{63.3} & \underline{65.8} & \textbf{68.0} & \textbf{70.1}\\
			7 & 70 & 63.1 & 65.5 & 67.3 & 69.4\\
			\hline
			\multicolumn{2}{c|}{$p$-value} & 0.4893 & 0.1842 & 0.0765 & 0.0874\\
		\end{tabular}
		\label{cutout(a)}
	}
	\quad\quad
	\subfloat[DeepLesion]{
		\begin{tabular}{cc|C{1cm}C{1cm}C{1cm}}
			\multirow{2}{*}{$n$} & \multirow{2}{*}{$s$} & \multicolumn{3}{c}{Image Size}\\
			& & 384 & 448 & 512 \\
			\hline\hline
			0 & 0 & 70.7 & 71.8 & 74.0 \\
			3 & 50 & 71.0 & 72.5 & 74.2 \\
			3 & 70 & 71.4 & \underline{72.8} & 74.4 \\
			5 & 50 & 71.3 & 72.6 & 74.1 \\
			5 & 70 & \textbf{71.9} & \textbf{73.2} & \underline{74.5} \\
			7 & 50 & \underline{71.6} & 72.7 & \textbf{74.7} \\
			7 & 70 & 71.5 & 72.6 & 74.3 \\
			\hline
			\multicolumn{2}{c|}{$p$-value} & 0.1445 & 0.0883 & 0.0981\\
		\end{tabular}
		\label{cutout(b)}
	}
	\caption{Investigation of number (denoted as $n$) and size (denoted as $s$) of masks in cutout. We simply adopt rectangle masks of side length $s$ in cutout. Note that in the above experiments, the baseline model ($n=0$ and $s=0$) corresponds to a vanilla CSD model. The best results are in bold while the second best are underlined. Note that we do not use adaptive cost function, noisy residual block or adversarial perturbation in these experiments. All p-values are calculated between the top-2 models in each column.}
	\label{cutout}
\end{table*}
\begin{table}[!t]
	\centering
	\begin{tabular}{c|ccc|c}
	Dataset & 1 & 3 & 5 & $p$-value\\
	\hline
	\hline
	DSB & 64.3 & 67.1 & 68.8 & 0.0379\\
	\hline
	DeepLesion & 72.8 & 75.0 & 76.5 & 0.0179\\
	\end{tabular}
	\caption{Influence of adding consistency regularization to different numbers of feature scales. \textbf{1,3} and \textbf{5} denote the number of feature scales. The default input sizes of DSB and DeepLesion are 448$\times$448 and 512$\times$512, respectively. All p-values are calculated between the top-2 models in each row.}
	\label{num_scale}
\end{table}
\begin{table}
	\small
	\centering
	\begin{tabular}{c|cccc|c}
	Dataset & SSD & \makecell[c]{Faster \\ RCNN} & \makecell[c]{Mask \\ RCNN} & \makecell[c]{RetinaNet \\ (ours)} & $p$-value\\
	\hline
	\hline
	DSB & 66.8 & 67.7 & 68.3 & 68.8 & 0.0438\\
	DeepLesion & 75.5 & 75.9 & 76.3 & 76.5 & 0.0379\\
	\end{tabular}
	\caption{The default input resolutions for DSB and DeepLesion are $448\times 448$ and $512\times 512$, respectively. All p-values are calculated between the top-2 models in each row. The labeled ratio is 50\%.}
	\label{backbones}
\end{table}
\begin{figure}[!t]
	\centering
	\includegraphics[width=1.0\columnwidth]{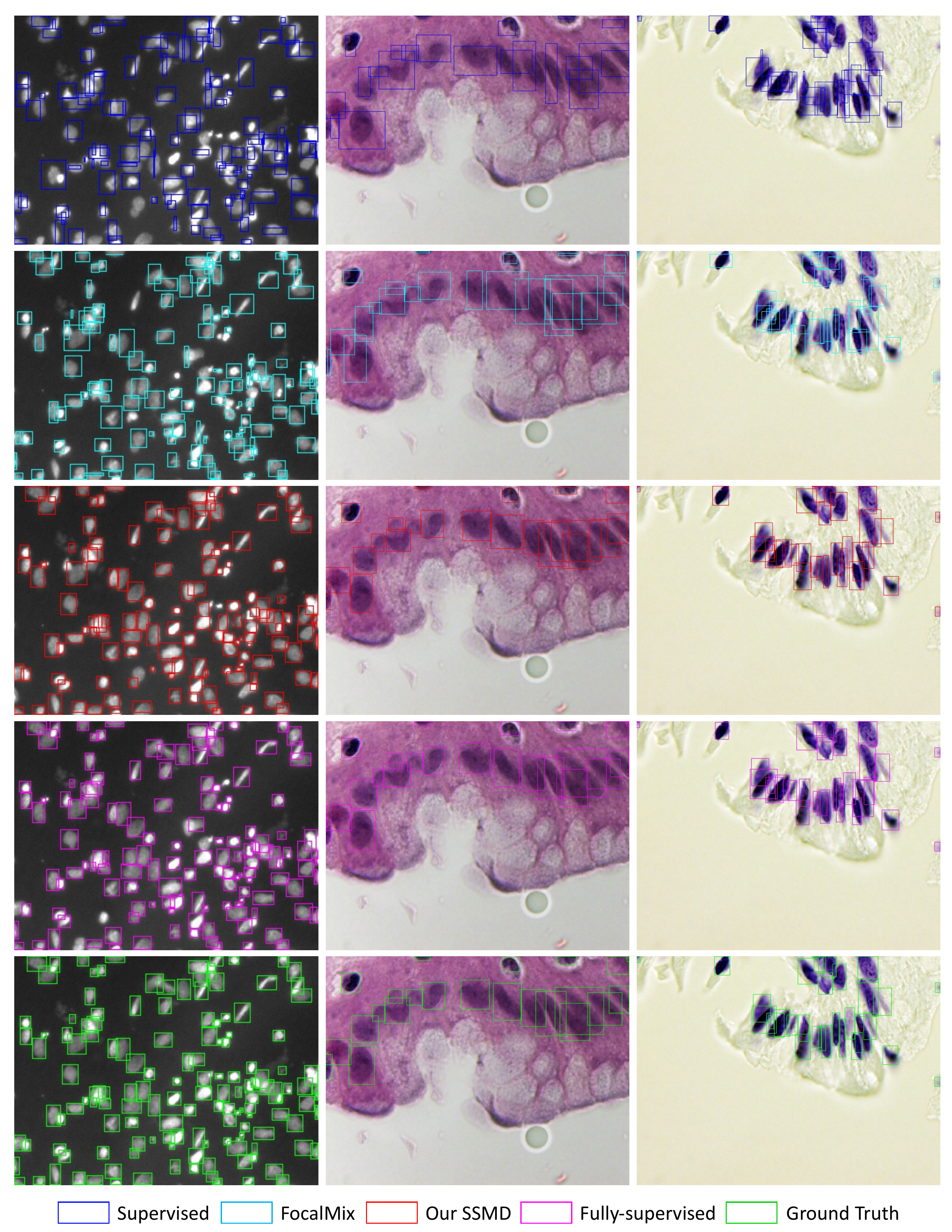}
	\caption{Visual comparison of different methods in DSB 2018. The colors of these approaches are consistent with those in Fig.3.}
	\label{vis_dsb}
\end{figure}
\begin{figure}[!t]
	\centering
	\includegraphics[width=1.0\columnwidth]{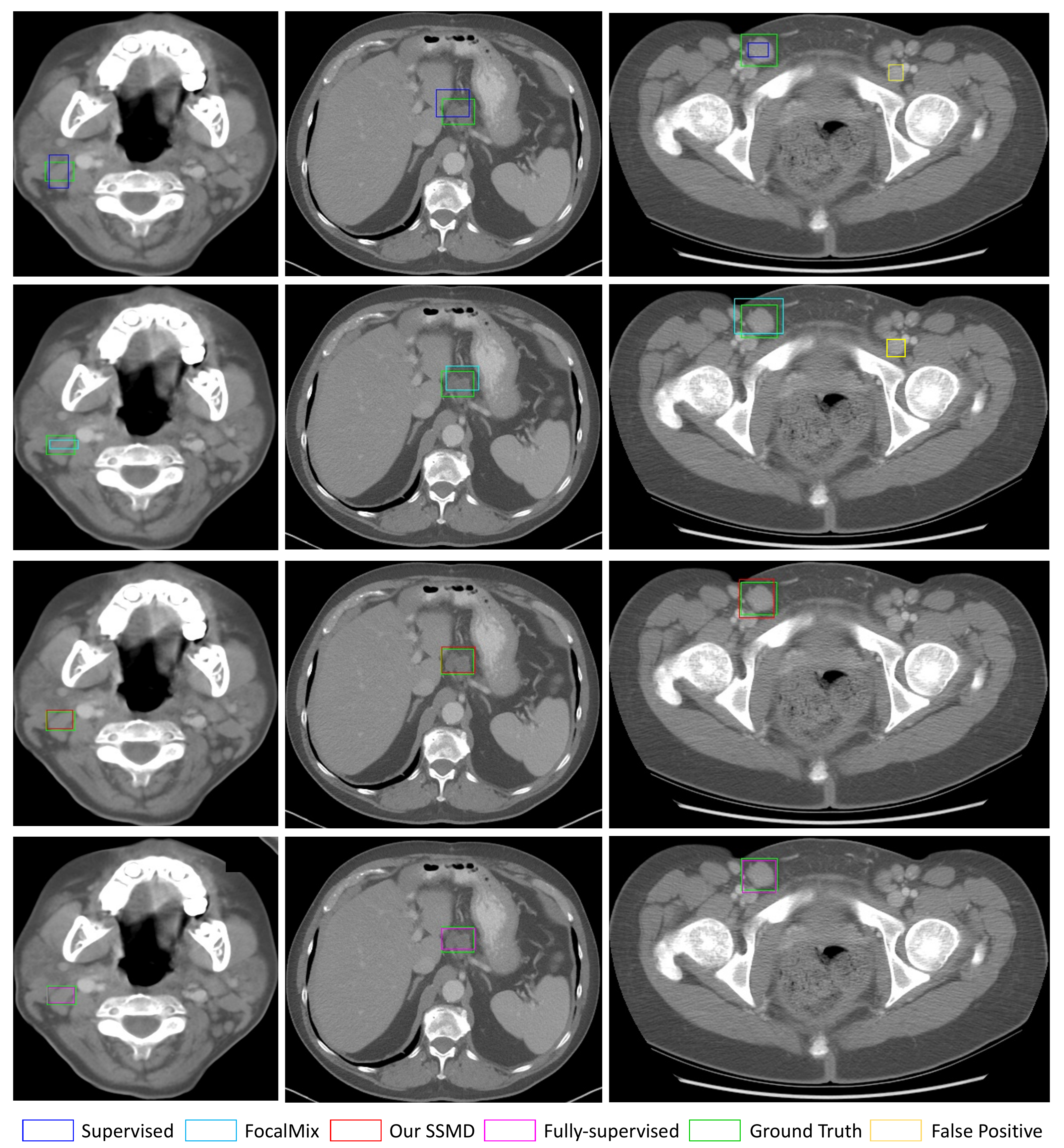}
	\caption{Visual comparison of different methods in DeepLesion. The colors of these approaches are consistent with those in Fig.3. Additionally, we use yellow boxes to denote the false positive results.}
	\label{vis_deeplesion}
\end{figure}
\begin{figure}[!t]
	\centering
	\includegraphics[width=1.0\columnwidth]{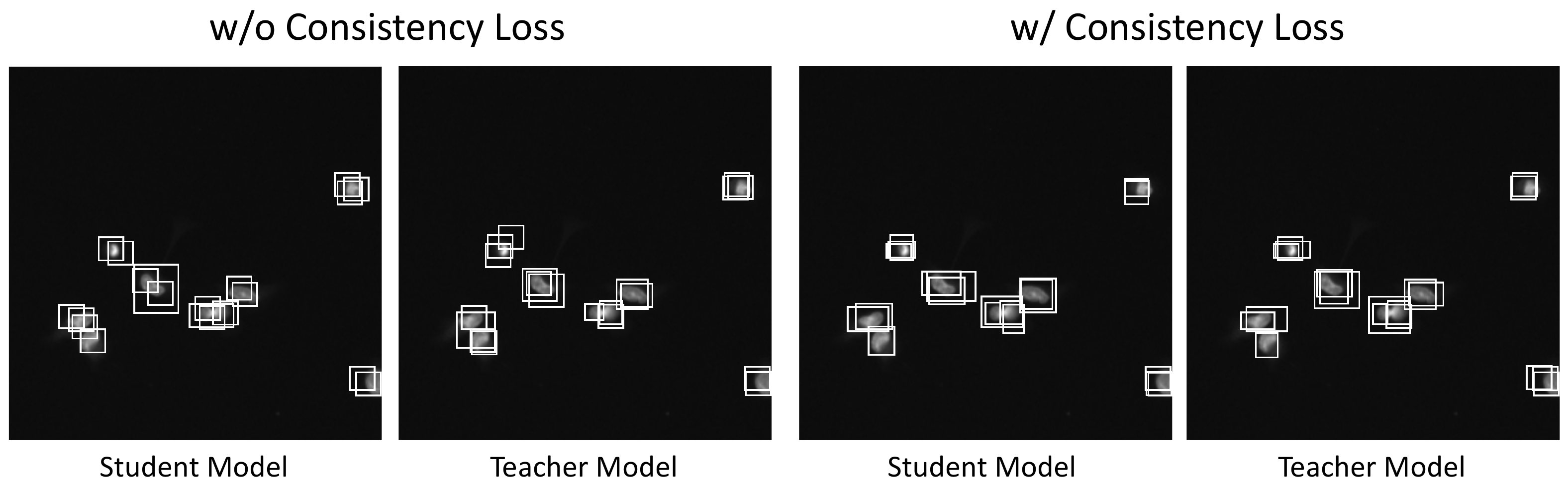}
	\caption{Proposal results from the student and the teacher model. For the left pair, we apply no consistency loss. For the right pair, we regularize the consistency of two models' predictions. Note that for each image, we display 20 proposals with highest confidence scores before the non-maximum suppression (NMS) step. \textbf{w/o} and \textbf{w/} stand for without and with. These results come from the detectors trained for 50 epochs.}
	\label{student_teacher}
\end{figure}
\textbf{Confidence Threshold $\tau$ } In Equation (\ref{adv_conf}), we introduce a confidence threshold $\tau$ in order to control the influence of various proposals to generated adversarial examples. We show the effects of employing different confidence thresholds in Fig.\ref{tau}. It can be seen that as $\tau$ increases, using adversarial perturbation produces better performance. However, a too large value may lead to a negative effect, such as 0.99. Such phenomenon is easy to explain since low-confidence regions might be incorrect and usually affect the detection accuracy negatively. In contrast, a threshold of 0.99 may filter out too many proposals. Based on the above observations, we set the default value of $\tau$ as 0.95 in both datasets.

\subsubsection{Investigation of EMA Factor}
Using EMA weights has been shown to help obtain more accurate predictions while stabilizing the training process \cite{tarvainen2017mean,zhou2020C2L}. Compared to CSD, we replace the siamese network with a student-teacher architecture to promote the detection performance.

In Table \ref{ema}, we gradually increase $\alpha$ in Equation (\ref{ema_formula}) from 0.8 to 0.999 and report its corresponding performance in both DSB and DeepLesion datasets. Setting $\alpha$ as 0.99 seems to be optimal in both datasets. In contrast, decreasing $\alpha$ may deteriorate the detection accuracy. Particularly, a small value such as 0.8 may be harmful to the consistency based approach, leading to worse performance than a vanilla CSD (65.2 in DSB and 74.0 in DeepLesion). The underlying reason may be that the teacher network is updated too frequently which makes its predictions unstable to some degree. Oppositely, a too large value (0.999) may be harmful since the teacher network is updated too sparsely. Considering these observations, we set $\alpha$ to 0.99 in the rest of the experiments.

\subsubsection{Investigation of Consistency Regularization}
In Table \ref{loss}, we present the performances of using different consistency regularization constraints. A simple baseline could be a model trained with normal and flipped images with manual annotations. Table \ref{loss} shows that adding consistency to predicted class scores improves the baseline. Specifically, simply regularizing the proposal confidence (denoted as $c$) can already bring more than 1 point improvement in both DSB and DeepLesion datasets. The largest improvements mainly come from the regularization of $x$ and $y$. In most cases, regularizing these two variables outperforms regularizing the variable $c$ only by approximate 1 point. The relative improvements in DeepLesion dataset are larger than those in DSB. Applying consistency regularization on all predicted variables brings about more than 3 points improvements over the supervised baseline. Compared with CSD, it can be observed that our proposed adaptive cost function leads to 1 point improvement.

\subsubsection{Number and Size of Masks in Cutout}
Cutout \cite{devries2017improved} is adopted in the proposed SSMD to add perturbations to input images. 
We apply cutout to the inputs of both student and teacher networks while adversarial perturbation is only applied to the input of teacher network.

As the number $n$ and the size $s$ (side length of a rectangle mask) of masks in cutout may affect the detection performance, it is necessary to conduct ablation studies for these two factors. In Table \ref{cutout}, the experimental results of different combinations of $n$ and $s$ are displayed. It is discovered that adding too many masks degrades the performance in both datasets. We argue that the reason can be summarized as: more masks usually means hiding more instances which make the detector fail to predict the correct boxes. In Table \ref{cutout} it is observed that as the input resolution increases, the effect of cutout becomes less significant. Specifically, $\{n$=5,s=$70\}$ and $\{n$=7, s=$50\}$ achieve two best and comparable results. $\{n$=7, s=$50\}$ seems to prefer larger input while $\{n$=5,s=$70\}$ affects more on small input resolutions. We take $\{n$=5,s=$70\}$ as the default setting in most experiments.

\subsubsection{{Adding Consistency Loss to Different Numbers of Scales}}
{To demonstrate the necessity of multi-scale consistency loss, we also report the experimental results of applying consistency loss to different numbers of scales in Table \ref{num_scale}. It suggests that reducing the number of applied scales would adversely affect the overall detection performance. Specifically, applying consistency loss to 3 scales makes SSMD achieve comparable results with FocalMix. Interesting, even if we further reduce this number to 1, SSMD still performs much better than the supervised baseline, demonstrating the effectiveness of proposed modules.}

\subsubsection{Investigation of Different Detection Backbones}

{In Table \ref{backbones}, we investigate the influence of using different detection backbones. As Table~\ref{backbones} shows, Mask RCNN \cite{he2017mask} achieves comparable results with our RetinaNet, although Mask RCNN runs much slower than our SSMD. Besides, Mask RCNN is better than Faster RCNN as it utilizes mask annotations during the training stage. SSD performs the worst across all 4 detection backbones, owing to its weakness in learning appropriate representations for medical image detection.}

\section{{Visualization}}
{In Fig.\ref{vis_dsb} and Fig.\ref{vis_deeplesion}, we visually compare the supervised model, FocalMix, our SSMD and the fully-supervised baseline to the ground truth in DSB 2018 and DeepLesion, respectively. These results again verify the effectiveness of proposed SSMD. As Fig.\ref{vis_dsb} and Fig.\ref{vis_deeplesion} display, SSMD is superior to FocalMix in two different aspects: precision and recall. In DSB, SSMD is able to predict more accurate bounding boxes for nucleus. In contrast, FocalMix sometimes misses small nucleus while SSMD has the ability to localize these hard cases and hence achieves higher recall. Moreover, SSMD is stronger on detecting severely overlapped nucleus, which demonstrates the strength of its learned powerful image representations. When we turn to DeepLesion, it is obvious that SSMD again produces more accurate box predictions. Besides, SSMD also shows its strength in reducing false positive predictions. When comparing SSMD with the fully-supervised baseline, we can see that SSMD achieves acceptable results in most cases.} {Besides the detection results, we also display the proposal results of the student and the teacher model in Fig.\ref{student_teacher}. The results indicate that the teacher models produce better detection results than those of student models in both pairs of images. If we compare their predictions carefully, it can be seen that the predictions of the student model with consistency loss are more accurate and consistent (with predictions of the teacher model) than those without consistency loss. These phenomena verify the findings from \cite{athiwaratkun2018there} which shows the teacher model helps to improve the learning process of the student model.}
\section{Conclusion}
In this paper, we proposed a novel semi-supervised medical detection method which can boost the fully-supervised performance with additional unlabeled data. Specifically, the proposed detector consists of an adaptive consistency cost function, noisy residual blocks and an instance-level adversarial perturbation strategy. We also conduct experiments which not only demonstrate the strength of the overall proposed detector on various settings, but also verify the effectiveness of each single proposed module. In the future, we will explore more ways to improve the proposed method.

\section*{Credit Author Statement}

\noindent \textbf{Hong-Yu Zhou:} Conceptualization; Investigation; Methodology; Software; Writing – original draft preparation \& revision.\\

\noindent \textbf{Chengdi Wang:} Investigation; Visualization; Validation; Writing - original draft preparation.\\

\noindent \textbf{Haofeng Li:} Investigation; Writing – original draft preparation.\\

\noindent \textbf{Gang Wang}: Investigation; Resources.\\ 

\noindent \textbf{Shu Zhang}: Investigation; Validation.\\ 

\noindent \textbf{Weimin Li}: Supervision; Writing - review \& editing.\\ 

\noindent \textbf{Yizhou Yu}: Conceptualization; Supervision; Writing - review \& editing; Funding acquisition.

\section*{Acknowledgement}
This work was funded in part by National Key Research and Development Program of China (No. 2019YFC0118101), National Natural Science Foundation of China (No. 91859203 and No. 82072005), Beijing Municipal Science and Technology Planning Project (No. Z211100003521009), Science and Technology Project of Chengdu (No. 2017-CY02–00030-GX), and Zhejiang Province Key Research \& Development Program (No. 2020C03073). This research was conducted while Hong-Yu Zhou was visiting West China Hospital, Sichuan University.
{
\clearpage
\bibliographystyle{ieee_fullname.bst}
\bibliography{egbib}
}
	
\end{document}